\documentclass{article}

\usepackage{arxiv}

\usepackage[utf8]{inputenc} 
\usepackage[T1]{fontenc}    
\usepackage{hyperref}       
\usepackage{url}            
\usepackage{booktabs}       
\usepackage{amsfonts}       
\usepackage{nicefrac}       
\usepackage{microtype}      
\usepackage{xcolor}         

\usepackage[american]{babel}

\usepackage{amsmath}
\usepackage{amssymb}
\usepackage{amsthm}
\usepackage{tikz}
\usetikzlibrary{arrows,cd,decorations.markings,shapes.geometric,shapes}
\usepackage{mathtools}
\usepackage{ebproof}

\usetikzlibrary{arrows,cd,decorations.markings,shapes.geometric,shapes}

\usepackage{tikzit}
\usepackage[all,cmtip]{xy}

\usepackage{tikzit}

\usetikzlibrary{circuits.ee.IEC}

\usetikzlibrary{calc,arrows.meta}

\newcommand\cofib\rightarrowtail

\newcommand\mdel[1]{}

\usepackage{txfonts}
\usepackage{pxfonts}

\makeatletter
\newcommand{\xdashrightarrow}[2][]{\ext@arrow 0359\rightarrowfill@@{#1}{#2}}
\newcommand*{\doublerightarrow}[2]{\mathrel{
  \settowidth{\@tempdima}{$\scriptstyle#1$}
  \settowidth{\@tempdimb}{$\scriptstyle#2$}
  \ifdim\@tempdimb>\@tempdima \@tempdima=\@tempdimb\fi
  \mathop{\vcenter{
    \offinterlineskip\ialign{\hbox to\dimexpr\@tempdima+1em{##}\cr
    \rightarrowfill\cr\noalign{\kern.5ex}
    \rightarrowfill\cr}}}\limits^{\!#1}_{\!#2}}}
\newcommand*{\triplerightarrow}[1]{\mathrel{
  \settowidth{\@tempdima}{$\scriptstyle#1$}
  \mathop{\vcenter{
    \offinterlineskip\ialign{\hbox to\dimexpr\@tempdima+1em{##}\cr
    \rightarrowfill\cr\noalign{\kern.5ex}
    \rightarrowfill\cr\noalign{\kern.5ex}
    \rightarrowfill\cr}}}\limits^{\!#1}}}
\makeatother

\makeatletter
\newcommand{\twoarrows}[3][0.2ex]{%
  \mathrel{\mathpalette\twoarrows@{{#1}{#2}{#3}}}%
}
\newcommand{\twoarrows@}[2]{\twoarrows@@#1#2}
\newcommand{\twoarrows@@}[4]{%
  \vcenter{\offinterlineskip\m@th
    \ialign{\hfil##\hfil\cr
      $#1#3$\cr
      \noalign{\vskip#2}
      $#1#4$\cr
    }%
  }%
}
\makeatother

\newcommand{\beq}{\begin{equation}}
\newcommand{\eeq}{\end{equation}}

\newcommand{\cop}{\mathsf{copy}}
\newcommand{\del}{\mathsf{del}}

\newtheorem{theorem}{Theorem}
\newtheorem{definition} {Definition} 
\newtheorem{assumption}{Assumption}

\usepackage[boxruled]{algorithm2e}
\usepackage{algorithmic}

\usepackage{multirow}
\usepackage{booktabs}
\usepackage{balance}
\usepackage{subfig}

\DeclareFontFamily{U}{dmjhira}{}
\DeclareFontShape{U}{dmjhira}{m}{n}{ <-> dmjhira }{}

\DeclareRobustCommand{\yo}{\text{\usefont{U}{dmjhira}{m}{n}\symbol{"48}}}

\usepackage{natbib} 
\usepackage{mathtools} 
\usepackage{booktabs} 
\usepackage{tikz} 

\title{Consciousness as a Functor \thanks{Draft under revision.} }

\author{ Sridhar Mahadevan \\
	Adobe Research and University of Massachusetts, Amherst\\
	\texttt{smahadev@adobe.com, mahadeva@umass.edu}
}

\hypersetup{
pdftitle={A template for the arxiv style},
pdfsubject={q-bio.NC, q-bio.QM},
pdfauthor={David S.~Hippocampus, Elias D.~Striatum},
pdfkeywords={First keyword, Second keyword, More},
}

\begin{document}
\maketitle

\begin{abstract}
As AI systems become progressively closer to achieving some type of AGI, the need for measuring their capabilities with respect to humans becomes more critical. Motivated by this criterion, we propose a novel theory of consciousness as a {\em functor} (CF) that receives and transmits contents from unconscious memory into conscious memory. Functors are similar to analogies such as ``the atom is like the solar system":  they not only  map objects in their domain category to corresponding objects in their co-domain category, they must also preserve the relationships among the objects in the domain category in the co-domain category.  Our CF framework can be seen as a categorial formulation of  Baars' Global Workspace Theory (GWT).  CF models the ensemble of unconscious processes as a topos category of coalgebras.  As every topos has an internal language defined by a Mitchell-B'enabou language with a Kripke-Joyal semantics, the internal ``language of thought" in CF is defined as a Multi-modal Universal Mitchell-B'enabou Language Embedding (MUMBLE).   We model the transmission of information from conscious short-term working memory to long-term unconscious memory using our recently proposed Universal Reinforcement Learning (URL) framework. To model the transmission of information from unconscious long-term memory into short-term memory, we propose a network economic model, where ``producer" agents correspond to unconscious processes, ``transporter" agents correspond to neural pathways from long-term to short-term memory, and ``consumer agents" correspond to short-term memory locations that use a competitive bidding process to manage the competition between unconscious long-term memory processes. Both URL and the network economic model of consciousness build on a formal theoretical framework for asynchronous parallel distributed computation without the need for synchronization by a global clock.  
\end{abstract}


\newpage 

\tableofcontents

\newpage 

\section{Introduction}\label{sec:intro}

Consciousness has been a topic of interdisciplinary study through the millennia. The term derives from the Latin word "conscius", a concatenation of "con" -- meaning "together" -- and  "scio" -- mearning "to know". The philosopher Ren\'e D\'escartes was one of the earliest to discuss consciousness at length in his book \citep{descartes}. The modern view arises from John Locke's definition of consciousness as ``The perception of what passes in a man's own mind" \citep{locke}. A popular geological metaphor in the 19th century attributed consciousness to hidden layers that ``recorded the past of an individual". In the late 20th century, consciousness became very actively studied with a large number of converging studies. A particularly influential theory was proposed by Baars \citep{baars:oup} called the Global Workspace Theory, which will be discussed in more detail in this paper.  Numerous book-length explanations of consciousness have been published in recent years, including philosophical treatises by \citet{dennett} and \citet{chalmers}, cognitive science theories \citep{baars:theater}, and neuroscientific investigations \citep{dehaene,Crick1990-CRITAN}.  \citet{edelman:consciousness} propose a Darwinian framework for modeling consciousness, which we will elaborate on in Section~\ref{network-economy-as-consciousness}. A popular account of a neuroscientific theory of Baars' Global Workspace Theory is given by \citet{dehaene}. There are now a number of academic journals that publish scholarly papers related to consciousness, including the Journal of Consciousness Studies. 

As AI systems are growing rapidly in their attempt to achieve some type of artificial general intelligence (AGI) \citep{agi-dm}, a computational theory of consciousness may become more useful in discerning human-like vs. machine-like cognition.  We propose a framework that models consciousness as a {\em functor} (CF), a structured mapping between categories \citep{maclane:71,riehl2017category} modeling conscious and unconscious processes. Our CF framework is part of a longer research project on the categorial foundations of AGI, which is now appearing as a book \citep{sm:aig}. Various parts of this research project have been published over the past few years, including universal imitation games \citep{sm:uig},  universal causality \citep{DBLP:journals/entropy/Mahadevan23,cktheory,mahadevan2025toposcausalmodels}, categorical models of generative AI architectures (GAIA) \citep{mahadevan2024gaiacategoricalfoundationsgenerative}, universal decision models \citep{sm:udm}, universal reinforcement learning (URL) \citep{mahadevan2025universalreinforcementlearningcoalgebras}, and categorical homotopy theory for generative AI and large language models (LLMs) \citep{mahadevan2025topostheorygenerativeai}. We refer the reader to some of these papers for additional examples of the modeling paradigm used in this paper, and we restrict our focus to only describing parts of the research program salient in modeling consciousness.

Following \citet{baars:oup}, we view conscious processing as highly sequential, deliberative, slow, and prone to errors. In contrast, unconscious processes are asynchronous, distributed, highly parallelized, and rapid. This dichotomy resembles the ``thinking slow, thinking fast" metaphor of \citet{kahneman2011thinking}. Our functorial framework can be contrasted with the more mechanistic model defined by \citet{blum2021theoreticalcomputerscienceperspective,blum:pnas} as the Conscious Turing Machine (CTM). To contrast our CF framework with CTM in terms of Figure~\ref{fig:three-layer}, our framework is at the level of a {\em computational theory}, in the sense defined by \citet{marr:personal-view}, whereas the CTM is more at the level of a particular algorithmic implementation. We note in passing that Turing machines can be easily modeled as a type of universal coalgebra \citep{jacobs:book,rutten2000universal}. A fundamental difference between CF and the CTM framework is that we do not assume the existence of any global clock signal to coordinate parallel processes. Our CF framework builds heavily on past work on asynchronous parallel distributed computation \citep{bertsekas:pdc,witsenhausen:1975}, which we believe is essential to theoretical models of the brain. 

CF is based on the use of universal coalgebras, which constitute a broad family of dynamical systems that are defined as $\alpha_F: X \rightarrow F(X)$, where $X$ is an object in some category ${\cal C}$, and $F$ is an endofunctor that specifies the dynamics. A simple and yet extremely general example of a coalgebra is defined by the {\em powerset} functor $\alpha_{\cal P}: X \rightarrow {\cal P}(X)$, which has been shown to admit a final coalgebra under some conditions \citep{aczel-final-coalgebra-thm}, where ${\cal P}(X)$ represents the powerset of the set $X$. To relate this definition to a simple automata model, note that we can define a nondeterministic finite state machine as mapping a given set $X$ of states to the powerset $X \times A$ of possible next states given a particular input token. Mealy and Moore machines can all be specified similarly, and we refer the reader to \citet{DBLP:conf/category/ArbibM74a} for a categorial formulation of automata theory. 

A large family  of probabilistic transition systems can also be formulated as coalgebras, and we refer the reader to \citep{SOKOLOVA20115095}. Of interest to us in AI are canonical models like Markov decision processes (MDPs) \citep{bertsekas:rlbook,sb:2018},  predictive state representations (PSRs) \citep{singh-uai04}, as well as deep learning models \citep{mahadevan2024gaiacategoricalfoundationsgenerative}. There is a rich theory of coalgebras that can be brought to bear on the problem of consciousness. In particular, we posit a {\em topos} category of coalgebras that defines the space of unconscious processes. A fundamental implication from recent work in categorical probability is that for the brain to neurally realize a wide spectrum of causal, probabilistic and statistical inference, it must have the capacity of copy, delete, and multiple objects \citep{Fritz_2020,DBLP:journals/mscs/JacobsKZ21}. This literature shows that it is possible to define reasoning under uncertainty as a {\em string diagram} \citep{Selinger_2010} in a symmetric monoidal category. Whilst this structure is sufficient to support reasoning under uncertainty, we posit that for consciousness, the additional structure provided by a topos of coalgebras allows defining an internal ``language of thought" \citep{fodor:loth} that arises due to the topos-specific structure. 

A {\em topos} \citep{maclane:sheaves,Johnstone:592033} is a category that has ``set-like" properties: it has all (co)limits (which in sets are essentially forming intersections and unions), admits a subobject classifier (a generalization of the notion of subset), and an exponential object for each pair of objects (which is the generalization of the property that the set of all functions from set $A$ to set $B$ is itself a set). The remarkable property of a topos is that it admits an internal language, a formal logic with a well-defined semantics. It is well-known that every topos has an internal language, a local set theory \citep{bell}, which has a formal Mitchell-B\'enabou Language (MBL) with a possible-worlds Kripke-Joyal semantics \citep{maclane:sheaves}. \citet{blum:pnas} propose an internal language for the CTM that they term ``Brainish". We show formally how in our CF framework,  such an internal language arises from the topos category of coalgebras that represent the ensemble of unconscious processors competing to place information in short-term memory. In particular, the category of coalgebras forms a {\em topos} \citep{maclane:sheaves}, a particular type of ``set-like" category that has all (co)limits, admits a subobject classifier and has exponential objects.  In our theoretical coalgebraic formulation of consciousness, ``Brainish" is defined as a Multi-modal Universal Language for Mitchell-B\'enabou Embeddings (MUMBLE). 

A fundamental problem in consciousness is modeling the flow of information between conscious and unconscious memory. To model the flow of information from highly deliberative, sequential, and error-prone short-term conscious memory into highly distributed, asynchronous, parallel long-term memory, we build on our  proposed framework of Universal Reinforcement Learning (URL) \citep{mahadevan2025universalreinforcementlearningcoalgebras},  which generalizes the standard reinforcement learning (RL) framework used to solve Markov Decision Processes (MDPs) to general universal coalgebras \citep{jacobs:book,rutten:streams,feys,kozen}. 
To model the reverse flow of information from highly parallel, asynchronous distributed unconscious processes into short-term conscious processes, we introduce the concept of modeling unconscious-to-conscious transmission as a {\em network economy} \citep{nagurney:vibook}, where ``producer agents" are unconscious processes that want to post their data into short-term memory locations, but must compete with other unconscious processes for the privilege to do so. The neural pathways from long-term unconscious memory to short-term memory are controlled by ``transporter agents", and the locations in short-term memory are managed by ``consumer agents" that use a competitive bidding process to determine who gets to post their data at their location. We show that both URL and the network economic model can be solved using the same asynchronous parallel distributed computational framework described in \citep{bertsekas:pdc,witsenhausen:1975}. 
Unlike the case of CTM,  in our CF framework, we do not posit the existence of any clock. Decisions are made in long-term memory asynchronously, and the intrinsic model and its categorical generalization, the Universal Decision Model (UDM) \citep{sm:udm} in our previous work shows how to make decisions asynchronously without assuming a global clock, as in CTM.  The asynchronous distributed computational framework has had much success in analyzing  the convergence of reinforcement learning methods, like Q-learning \citep{tsitsiklis}, which draws upon the basic theory of parallel and distributed computation developed in \citep{bertsekas:pdc}. 

To summarize the goals of this paper, we provide a high-level introduction to our CF framework, relegating most of the details to our forthcoming book \citep{sm:aig} and our previously published papers on the specific components of our framework. Here is a brief roadmap to the rest of the paper. In Section~\ref{marr-consciousness}, we introduce the notion of a computational theory of consciousness, drawing upon Marr's advocacy of a three-level modeling hierarchy \citep{marr:personal-view}.  In Section~\ref{cf-arch}, we describe a high-level diagram of our CF architecture.   In Section~\ref{introcat-cf}, we introduce formally the mathematical machinery of categories and functors for modeling consciousness. In Section~\ref{topos-url}, we introduce the use of topos categories for modeling consciousness, and illustrate a simple example showing that the category of (action)value functions in RL forms a topos. In Section~\ref{mumble}, we introduce the internal language of thought in our CF framework defined by the Multimodal Universal Mitchell-B\'enabou Language Embedding (MUMBLE). We define the language formally, and give its Kripke-Joyal semantics. We define a special case of the language for the topos category of sheaves, which are constructed from Yoneda embeddings. This case has particular importance in applications, such as universal causality \citep{mahadevan2025toposcausalmodels} as well as in language models \citep{mahadevan2025topostheorygenerativeai}. Section~\ref{udm} introduces the framework of asynchronous parallel distributed computation \citep{bertsekas:pdc,witsenhausen:1975}, which we adapt to the categorial setting of the topos of coalgebras.  Section~\ref{network-economy-as-consciousness} introduces the idea of modeling the transmission of information from the large pool of unconscious processes into the resource-limited short-term memory locations as a network economy comprising of three tiers of agents, and we introduce the formal mathematics of solving variational inequalities \citep{nagurney:vibook}, and present a stochastic approximation method for solving VI's that is both asynchronous and highly decentralized. Finally, we summarize the paper in Section~\ref{summary} and outline briefly some directions for future work. 

\section{Towards A Computational Theory of Consciousness}
\label{marr-consciousness} 
\begin{figure}[t]
    \centering
      \includegraphics[width=0.75\linewidth]{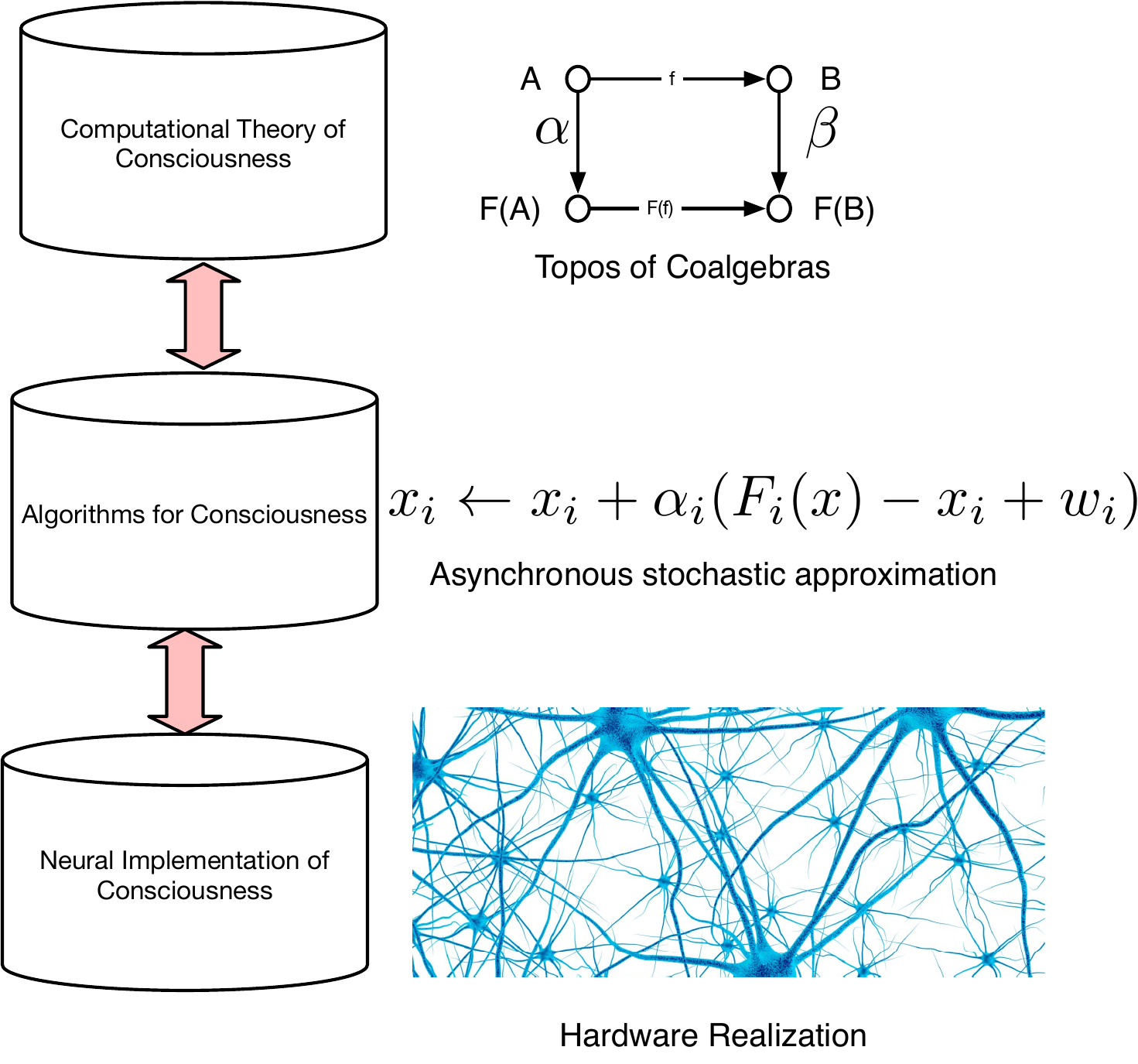}
    \caption{Consciousness can be modeled at three layers, an approach advocated by David Marr \citep{marr:personal-view}.}
    \label{fig:three-layer}
\end{figure}

 We are influenced by the philosophy of AI pioneer David Marr \citep{marr:personal-view,Marr:1982:VCI:1095712}, who argued that any complex information processing system, such as the brain, has to understood at multiple levels, and he paid particular emphasis to the top {\em computational theory} level, over the middle {\em algorithmic layer} and the bottom {\em neural implementation} layer. Figure~\ref{fig:three-layer} illustrates David Marr's paradigm applied to the study of consciousness. Our paper can be viewed as primarily addressing the top  layer of developing a computational theory of consciousness, and the algorithmic structure in the middle layer. A paradigmatic example is the computational theory of reinforcement learning \citep{sb:2018}, where the top layer is the theory of Markov decision processes \citep{PUTERMAN1990331}, the middle algorithmic layer is defined by algorithms, such as TD-learning \citep{sb:2018}, and the bottom neural implementation layer refers to the realization of TD using dopamine neurotransmitters (the neural instantiation of RL is described in detail in \citep{sb:2018}). 
 
 This decomposition helps us distinguish our work from the vast literature in the field. 
As alluded to in Section~\ref{sec:intro}, the literature on consciousness is enormous, spanning many fields of inquiry and over several centuries. Many books have been published on the topic, just in the last few decades, seom from the standpoint of neuroscience \citep{Crick1990-CRITAN,dehaene}, some from the standpoint of cognitive science \citep{baars:oup,baars:theater,chalmers,chalmers:theory},  and some from the standpoint of philosophy \citep{searle:consciousness,dennett}.  None of these expositions really accounts for a computational theory of consciousness. 

\citet{marr:personal-view} explained the difference between what he termed a {\em Type I} theory from a {\em Type II} theory. The former is exemplified by Chomsky's work on universal grammar \citep{chomsky1965} and formal language theory \citep{chomsky59}, as well as the theory of optimal sequential decision making \citep{DBLP:books/lib/Bertsekas05}. In these examples, the goal is to articulate a theory of language or optimal behavior, but no effort is made to explain its neural substrate. A Type II theory, on the other hand, might be an explanation of how proteins fold, which is the result of a large number of forces that act on a protein molecule. We are of course unsure whether consciousness can be explained by a Type I or Type II theory, but a general guide here is that if a cognitive phenomenon is universal -- we are all conscious -- and not the result of some elaborate process of learning or training over decades, like getting a Ph.D., it is likely that there exists a Type I theory of that phenomenon. 

\citet{Crick1990-CRITAN} make some general observations about consciousness that are worth emphasizing, since they pertain to the general theme of our work, even if unlike them, we are proposing a far more abstract computational theory. They make the following general points, which are worth highlighting: 

\begin{itemize}
    \item We all understand at an intuitive level what it means to be conscious, and it is probably premature at this stage to demand a precise definition of consciousness. However, we believe it is worthwhile to develop a computational theory of consciousness, as it may be useful to posit theoretical entities that play a role in realizing consciousness. 

    \item The language system may play a minor role in consciousness, as it is likely to be a more general feature of animal brains, rather than purely human brains. Accordingly, we do not ascribe any particular language aspects to consciousness. However, following \citet{fodor:loth}, we propose a ``language of thought", which we define as MUMBLE (Multimodal Universal Mitchell-B'enabou Embedding). MUMBLE can be viewed as a categorial version of Brainish in CTM. 

    \item There may be many forms of consciousness associated with perception, thought, emotion, pain etc. This multimodal nature of consciousness will be reflected in the design of our Multimodal Universal Mitchell B\'enabou Language Embedding (MUMBLE). 

    \item In the work on philosophy of consciousness \citep{searle:consciousness}, there is a great emphasis on {\em qualia}, which is that the sensation of red that I see when looking at a STOP traffic sign in the U.S. may be different from the sensation of red that others see. This subjectivity implies that consciousness is not amenable to scientific study. We do not get into such debates, preferring to hypothesize that as all of us are conscious, there is an objective scientific problem that demands an explanation. 
\end{itemize}

One can ask what should be required of a computational theory of consciousness. Some posit that consciousness can be modeled probabilistically as low-dimensional projection of a joint distribution  \citep{bengio2019consciousnessprior}. We prefer to model consciousness more abstractly in terms of categorical concepts, which can easily be reducible to statements about distributions. For example, Markov categories \citep{Fritz_2020} provide an elegant string diagrammatic category theory language for causal, probabilistic and statistical reasoning in a symmetric monoidal category. Our treatment makes a ``least commitment" philosophy of not burdening our theoretical machinery with particular concrete categories, unless it is absolutely essential. For example, we emphasize the identification of universal properties that underlie consciousness, following the work we have done in the past on universal causality \citep{DBLP:journals/entropy/Mahadevan23}, universal decision models \cite{sm:udm}, and more recently, universal reinforcement learning \citep{mahadevan2025universalreinforcementlearningcoalgebras}. In each of these cases, our goal was to identify what can be stated about the cognitive phenomena in question -- be it causality, decision making or learning -- in terms of categorical concepts. 

A highlight of our approach to consciousness is the investigation of the question: can we identify universal properties that underlie consciousness? In category theory, a property is universal if it can be defined in terms of an initial or final object in a category of diagrams \citep{riehl2017category}, or in terms of a {\em representable} functor. In defining an architecture for consciousness as a functor, we place a special emphasis on understanding the universal properties of such diagrams. To make this more concrete, if we view consciousness at a broad level as involving the transmission of information between a very large number of unconscious processes, which constitute ``long-term" memory, into a relatively small in comparison ``short-term" memory, we can model this information transmission categorically in terms of functors that map between the category of unconscious processes to the category of conscious elements. We can inquire as to the universal properties of the two categories in question, as well as that of the functors involved. 

\section{A Birds Eye View of Our Consciousness Framework} 
\label{cf-arch} 

\begin{figure}[t]
    \centering
    \includegraphics[width=0.75\linewidth]{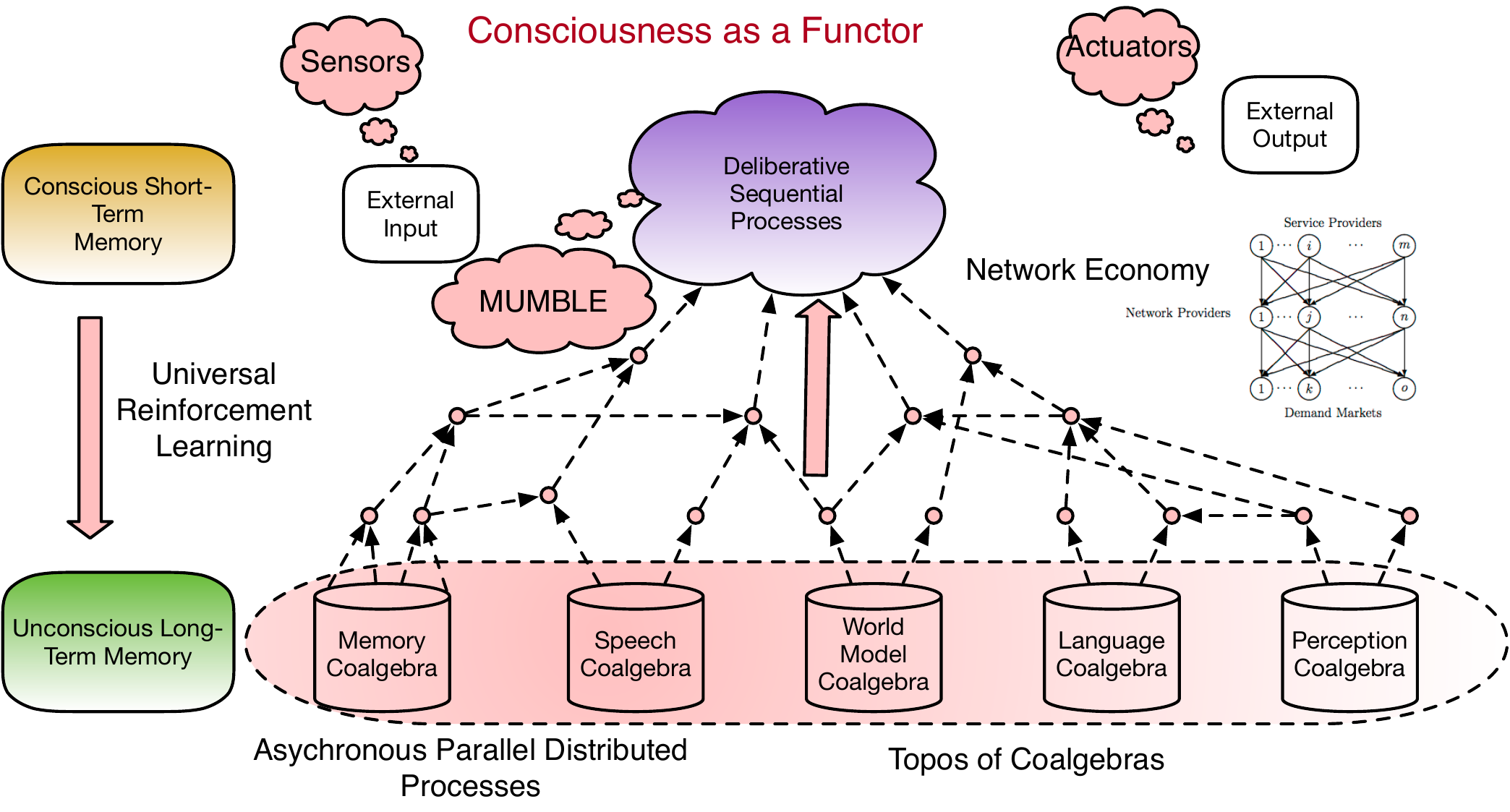}
    \caption{Architecture of our Consciousness as a Functor (CF).}
    \label{fig:arch}
\end{figure}

We begin with a high-level pictorial illustration of our consciousness framework, building on the past insights of Baars' Global Workspace Theory \citep{baars:oup}, and the Conscious Turing Machine \citep{blum:pnas}. Figure~\ref{fig:arch} illustrates the high-level architecture of our consciousness framework, whose main components we describe below. 

\begin{enumerate}
    \item {\em Unconscious processes as coalgebras:} Each unconscious process in long-term memory is modeled as a coalgebra $\alpha_F: X \rightarrow F(X)$, whose $F$ dynamics is specified as a functor. 

    \item {\em Topos of unconscious process coalgebras:} The ensemble of coalgebras defining the unconscious processes defines a topos category, which is closed under the operation of taking (co)limits, admits a subobject classifier, and has exponential objects These properties are akin to saying the topos category is ``set-like", in that it allows all the common operations one can do with sets, generalized to processes. 

    \item {\em Diagram functor modeling ``up-tree" competition architecture:} The ``up-tree" competition binary tree of the CTM is generalized into an arbitrary functor diagram, which maps from the topos category of unconscious processes into the category of conscious short-term memory. The diagram may have a far more complex structure than a tree, and this flexibility is exploited in our framework. 

    \item {\em Universal Reinforcement Learning}: Our recently proposed framework of URL \citep{mahadevan2025universalreinforcementlearningcoalgebras} is used in our consciousness framework to manage the asynchronous decentralized parallel computation among the unconscious processes that are competing to place their outputs in short-term memory. URL generalizes RL from the solving of MDPs or related dynamical systems to finding final coalgebras in universal algebras. 

    \item {\em MUMBLE:} Exploiting the property that the unconscious processes are defined as a topos of coalgebras, which admit a formal internal language, we define the internal language of the mind as MUMBLE, or Multi-Modal Universal Mitchell-B'enabou Language Embedding. MUMBLE is defined formally in this paper, and we specify its Kripke-Joyal semantics. MUMBLE can be seen as a formalization of Brainish in the CTM. 

    \item {\em Network economic model of information transmission from unconscious long-term memory into short-term memory}: A fundamental contribution of our CF framework is to introduce the idea of modeling transmission of information into the resource-limited short-term memory as a problem in network economics \citep{nagurney:vibook}, in contrast to other approaches, such as ``up-tree" competition \citep{blum:pnas} or dimensionality reduction \citep{bengio2019consciousnessprior}. We introduce the formal variational inequality (VI) formalism for solving network economies, and describe an asynchronous parallel distributed method for solving VIs that can work without the need for global coordination signals. 
\end{enumerate}

In the remainder of the paper, we elaborate on all of these components of our architecture, and study their formal properties. 

\section{Categories, Functors, and Coalgebras for Modeling Consciousness} 
\label{introcat-cf} 

\begin{figure}
    \centering
    \includegraphics[width=0.5\linewidth]{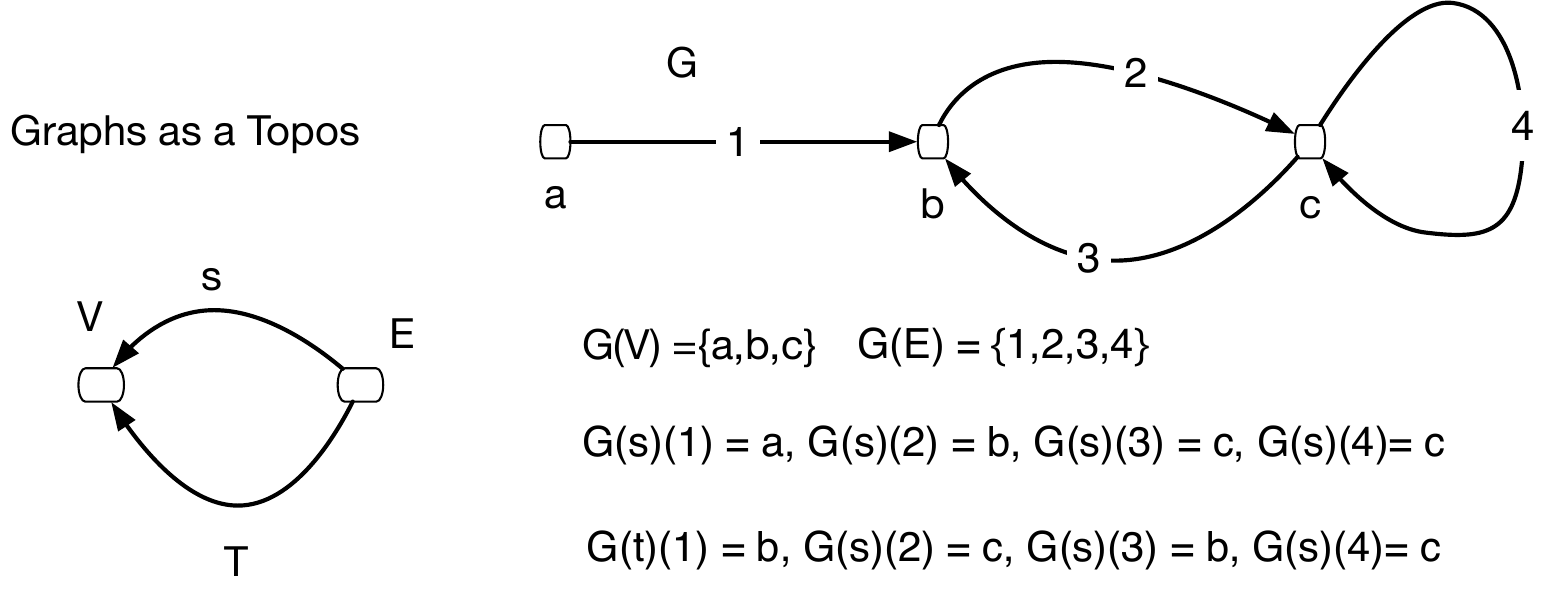}
    \caption{Category theory is the study of universal properties. While directed graphs have been widely studied for centuries, the realization that the category of graphs forms a topos reveals unexpected insights into their compositional properties, like how to construct the exponential of two graphs, or a subobject classifier  \citep{vigna2003guided}.}
    \label{fig:topos-of-graphs}
\end{figure}

Since we anticipate the reader as being unfamiliar with the theory of categories and functors, we provide a brief review here, and recommend standard treatments for the missing details \citep{maclane:71,riehl2017category}. Categories generalize other parts of  mathematics with which the reader may be familiar, such as algebra, calculus, graph theory, optimization, probability,  statistics, or topology. All of these subfields emerge as special types of categories with different universal properties. 

A category ${\cal C}$ is essentially a collection of abstract {\em objects} $c \in {\cal C}$. Anything technically can count as an object, from a set, a measurable space, or a Turing machine. Each category ${\cal C}$ is additionally specified by a set of arrows ${\cal C}(c,d)$ between each pair of objects $c$ and $d$. There is an identity arrow $1_c \in {\cal C}(c,c)$. Arrows compose in the obvious way, inducing a function ${\cal C}(c,d) \times {\cal C}(d,e) \rightarrow {\cal C}(c,e)$.   

There is an immense variety of ways that categories can be specified, and there is a wealth of examples in applied category theory textbooks \citep{fong2018seven}. In addition, category has been used to develop state-of-the-art data visualization methods, such as UMAP \cite{umap}, and resolve an impossibility theorem in data clustering \citep{Carlsson2010}. Relational databases are storehouses of knowledge widely used in industry and can be formally shown to define categories \citep{Spivak_2012}. 

A {\em functor} $F: {\cal C} \rightarrow {\cal D}$ between two categories ${\cal C}$ and ${\cal D}$ is specified by an {\em object function} mapping each $c \in {\cal C}$ to $Fc \in {\cal D}$, and an {\em arrows function} mapping each arrow $f \in {\cal C}(c,d)$ to $Ff \in {\cal D}(Fc, Fd)$.  Functors come in many varieties, but two important cases are worth highlighting. 

\begin{definition}  
A {\bf {covariant functor}} $F: {\cal C} \rightarrow {\cal D}$ from category ${\cal C}$ to category ${\cal D}$, is defined as \mbox{the following: }

\begin{itemize} 
    \item An object $F X$ (also written as $F(X)$) of the category ${\cal D}$ for each object $X$ in category ${\cal C}$.
    \item An  arrow  $F(f): F X \rightarrow F Y$ in category ${\cal D}$ for every arrow  $f: X \rightarrow Y$ in category ${\cal C}$. 
   \item The preservation of identity and composition: $F \ id_X = id_{F X}$ and $(F f) (F g) = F(g \circ f)$ for any composable arrows $f: X \rightarrow Y, g: Y \rightarrow Z$. 
\end{itemize}
\end{definition}  

\begin{definition}  
A {\bf {contravariant functor}} $F: {\cal C} \rightarrow {\cal D}$ from category ${\cal C}$ to category ${\cal D}$ is defined exactly like the covariant functor, except all the arrows are reversed. 
\end{definition}  

In modeling computation by coalgebras, we assume categories that are {\em concrete}. 

\begin{definition}
    A functor $F: {\cal C} \rightarrow {\cal D}$ is {\em full} if the assignment 
    \[ F: {\cal C}(c, c') \rightarrow {\cal D}(F(c), F(c')) \]
    is surjective. The functor $F$ is {\em faithful} if the above assignment is injective. A functor is fully faithful if the above assignment is a bijection. 
\end{definition}

\begin{definition}
    A category ${\cal C}$ is {\em concrete} if it admits a faithful set-valued functor $F: {\cal C} \rightarrow {\bf Sets}$. 
\end{definition}

To illustrate the application of functors, Figure~\ref{fig:topos-of-graphs} shows that any directed graph can be modeled categorically as a contravariant functor from a category with just two objects -- one denoting an abstract ``vertex" object and the other an abstract ``edge" object -- into the category of sets.  Each graph $G$ is modeled as a contravariant functor from a category with two objects $V$ and $E$ and two non-identity arrows $s$ and $t$. A particular graph, such as $G$ shown, is then defined as a functor that maps each object -- $V$ and $E$ here -- into a set, and each arrow -- $s$ and $t$ here -- into a function on sets. The structure of such functors can be shown to form a topos \citep{vigna2003guided}, which is a ``set-like" category with special properties. We will see that the category of unconscious processes will also be modeled as a topos, and discuss its universal properties later in the paper. 

\subsection{Labeled Transition Systems as Coalgebras} 

Coalgebras \citep{jacobs:book,rutten2000universal} play a central role in our theory of consciousness, as we model each unconscious process as a coalgebra.  Coalgebras are a categorical framework for labeled transition systems, which covers a vast range of  dynamical systems, from finite state automata, grammars and Turing machines, to stochastic dynamical systems like Markov chains,  MDPs or PSRs \citep{jacobs:book,rutten2000universal,SOKOLOVA20115095}.  A coalgebra is simply defined as the structure 

\[ \alpha_F =  X \rightarrow F(X) \]

where $X$ is an object in some category ${\cal C}$, usually referred to as the {\em carrier}, and the functor $F$ defines the $F$-dynamics of the coalgebra. As an example of a coalgebra, consider the functor ${\cal P}_X: X \rightarrow 2^X$ that maps from a set $X$ to its powerset $2^X$ in the category ${\bf Sets}$ of sets. 

\begin{definition} 
    A {\bf labeled transition system} (LTS) $(S, \rightarrow_S, A)$ is defined by a set $S$ of states, a transition relation $\rightarrow_S \subseteq S \times A \times S$, and a set $A$ of labels (or equivalently, ``inputs" or ``actions"). We can define the transition from state $s$ to $s'$ under input $a$ by the transition diagram $s \xrightarrow[]{a} s'$, which is equivalent to writing $\langle s, a,  s' \rangle \in \rightarrow_S$. The ${\cal F}$-coalgebra for an LTS is defined by the functor 

    \[ {\cal F}(X) = {\cal P}(A \times X) = \{V | V \subseteq A \times X\} \]
\end{definition} 
We can also define a category of $F$-coalgebras over any category ${\cal C}$, where each object is a coalgebra, and the morphism between two coalgebras is defined as follows, where $f: A \rightarrow B$ is any morphism in the category ${\cal C}$. 
\begin{definition} 
Let $F: {\cal C} \rightarrow {\cal C}$ be an endofunctor. A {\em homomorphism} of $F$-coalgebras $(A, \alpha)$ and $(B, \beta)$ is an arrow $f: A \rightarrow B$ in the category ${\cal C}$ such that the following diagram commutes:

\begin{center}
\begin{tikzcd}
  A \arrow[r, "f"] \arrow[d, "\alpha"]
    & B \arrow[d, "\beta" ] \\
  F(A) \arrow[r,  "F(f)"]
& F(B)
\end{tikzcd}
\end{center}
\end{definition} 

For example, consider two labeled transition systems $(S, A, \rightarrow_S)$ and $(T, A, \rightarrow_T)$ over the same input set $A$, which are defined by the coalgebras $(S, \alpha_S)$ and $(T, \alpha_T)$, respectively. An $F$-homomorphism $f: (S, \alpha_S) \rightarrow (T, \alpha_T)$ is a function $f: S \rightarrow T$ such that $F(f) \circ \alpha_S  = \alpha_T \circ f$. Intuitively, the meaning of a homomorphism between two labeled transition systems means that: 
\begin{itemize}
    \item For all $s' \in S$, for any transition $s \xrightarrow[]{a}_S s'$ in the first system $(S, \alpha_S)$, there must be a corresponding transition in the second system $f(s) \xrightarrow[]{a}_T f(s;)$ in the second system. 

    \item Conversely, for all $t \in T$, for any transition $t \xrightarrow[]{a}_T t'$ in the second system, there exists two states $s, s' \in S$ such that $f(s) = t, f(t) = t'$ such that $s \xrightarrow[]{a}_S s'$ in the first system. 
\end{itemize}
If we have an $F$-homomorphism $f: S \rightarrow T$ with an inverse $f^{-1}: T \rightarrow S$ that is also a $F$-homomorphism, then the two systems $S \simeq T$ are isomorphic. If the mapping $f$ is {\em injective}, we have a  {\em monomorphism}. Finally, if the mapping $f$ is a surjection, we have an {\em epimorphism}. 

Turing machines \citep{arora} are a well-established theoretical model of computation, which were used in the Conscious Turing Machine model by \citet{blum:pnas}. We briefly describe how to model Turing machines as coalgebras, and further details can be found in \citet{jacobs:book}. To fix some notation, in coalgebraic specifications, $X^n = X \times X \times \cdot X$ denote the categorical product, and $n \cdot X = X + X + \cdot + X$ denotes the categorical co-product. A Turing machine has two types of states: {\em steering} states refer to the states in the finite-state controller, and {\em register} states refer to the contents of the tape that is read by the Turing machine. If the Turing machine has $n$ steering states, it can be specified as the coalgebra 

\[ S \longrightarrow \left(F(n \cdot S) \right)^n\]

which can be seen as a separate coalgebra $S \rightarrow F(n \cdot S)$ specifying the machine's behavior for each control state. For a concrete example of how to specify a simple Turing machine in this formalism, we refer the reader to Section 2.2.6 in \citet{jacobs:book}. 

\subsection{Stochastic Coalgebras}

Coalgebras can model stochastic systems as well, which is of significant interest in modeling consciousness.  \citet{SOKOLOVA20115095} describes how to build an entire language of {\em probabilistic} coalgebras, using the context-free grammar: 

\[ F \coloneqq \mbox{\textunderscore}  \ | \ A \ | \ \mbox{\textunderscore}  ^A  \ | \ {\cal P} \ | \ {\cal D} \  | \ F \circ F \ | \ F \times F \ | \ F + F \]

where $F$ denotes a functor constructed from this grammar. Here, $\mbox{\textunderscore}$ is the identity functor over the category {\bf Sets}. $A$ is the constant functor mapping any set to the fixed set $A$. ${\bf id}^A$ defines a mapping from any set $X$ to the set of all functions from $A$ to $X$. The {\em powerset} functor ${\cal P}$ maps a set $X$ to its collection of subsets. Most importantly, the probability distribution functor ${\cal D}$ is defined as follows. 

\begin{definition} 
    The {\bf probability distribution functor} ${\cal D}$ is defined as ${\cal D}: {\bf Sets} \rightarrow {\bf Sets}$ maps a set $X$ to ${\cal D} X = \{ \mu: X \rightarrow \mathbb{R}^{\geq 0} | \mu[X] = 1 \}$, and a function $f: X \rightarrow Y$ to ${\cal D}f: {\cal D} X \rightarrow {\cal D} Y$ as $({\cal D} f)(\mu) = \lambda.\mu[f^{-1}(\{y \})$. 
\end{definition} 

In plain English, a distribution functor constructs a probability distribution over any set that has finite support, and given any function $f$ from set $X$ to set $Y$, maps any element $y$ in the codomain ${\cal D} Y$ to the probability mass assigned by to its preimage by ${\cal D}f$.   We can now define an entire family of stochastic coalgebras as shown in Table~\ref{stochcoalg}. We follow the terminology introduced in \citep{SOKOLOVA20115095}. To translate into the RL language, Segala systems correspond to MDPs, and Vardi systems are essentially concurrent Markov chains. It is possible to derive properties of the entire family of such coalgebras using the universal properties sketched out in the previous section, which we will refer the reader to \citep{SOKOLOVA20115095}. Our goal here is to adapt this categorical language into the URL framework, which we discuss next. 

\begin{table}[h]
    \centering
    \begin{tabular}{|c|c|c|} \hline 
         \bf{Coalg}$_F$ & $F$ & {\bf Explanation}  \\ \hline 
         {\bf MC} & {\cal D} & Markov chain \\ \hline 
        {\bf DLTS} & $(\mbox{\textunderscore} + 1)^A$ & Deterministic automata \\ \hline 
        {\bf LTS} & ${\cal P}(A \times \mbox{\textunderscore} ) \simeq {\cal P}^A $ & Non-deterministic automata  \\ \hline 
        {\bf React} &  $({\cal D} + 1)^A$ & Reactive systems \\ \hline 
        {\bf Generative} & ${\cal D}(A \times \mbox{\textunderscore}) + 1$ & Generative Systems \\ \hline 
        {\bf Str} &  ${\cal D} + (A \times \mbox{\textunderscore}) + 1$ & Stratified systems \\ \hline 
        {\bf Alt} & ${\cal D} + {\cal P}(A \times \mbox{\textunderscore}) $ & Alternating systems \\ \hline 
        {\bf Var} & ${\cal D}(A \times \mbox{\textunderscore}) + {\cal P}(A \times \mbox{\textunderscore}) $ & Vardi systems \\ \hline 
        {\bf SSeg} & ${\cal P}(A \times {\cal D}) $ & Simple Segala Systems \\ \hline 
        {\bf Seg} & ${\cal P} {\cal D}(A \times \mbox{\textunderscore})$ & Segala systems \\ \hline 
        {\bf Bun} & ${\cal D} {\cal P}(A \times \mbox{\textunderscore})$ & Bundle systems \\ \hline 
        {\bf PZ} & ${\cal P} {\cal D} {\cal P} (A \times \mbox{\textunderscore})$ & Pneuli-Zuck systems \\ \hline 
        {\bf MG} & ${\cal P} {\cal D} {\cal P} (A \times \mbox{\textunderscore} \times \mbox{\textunderscore})$ & Most general systems \\ \hline 
    \end{tabular}
    \caption{Stochastic Coalgebras in URL using the notation from  \citep{SOKOLOVA20115095}.}
    \label{stochcoalg}
\end{table}

\section{Topos Theory for Modeling Consciousness}
\label{topos-url} 

In this and the next section, we introduce more formally the application of topos theory \citep{maclane:sheaves,Johnstone:592033,bell,goldblatt:topos} to modeling consciousness. As illustrated in Figure~\ref{fig:arch}, our CF framework assumes that unconscious processes are modeled as coalgebras, a categorial language for dynamical systems \citep{rutten2000universal}. \citet{rutten:streams} defines a coinductive calculus for defining {\em streams} of data, which seems particularly apt in modeling consciousness as a stream of data flowing between unconscious long-term memory and conscious short-term memory. We want to build the CF framework out of these unconscious coalgebraic processes by assembling them into a coalgebraic topos.

If we posit that URL maps deliberative sequential behavior in short-term memory into asynchronous distributed unconscious processes in long-term memory, then the resulting value functions associated with those processes can be shown to form a topos. We briefly summarize the result shown in our earlier paper \citep{mahadevan2025universalreinforcementlearningcoalgebras} on URL that the category of action-value functions ${\cal C_Q}$ forms a topos, and refer the reader to that paper for details. 

\subsection{Category of Action-Value Functions forms a Topos}

We can show this result by showing that the category ${\cal C_Q}$ is (co)complete, meaning all (co)limits exist, but that they have other properties that make them into a category called a topos \citep{maclane:sheaves,Johnstone:topostheory} that is a ``set-like" category with very special properties, which we will explore in the rest of the paper.  A topos generalizes all common operations on sets. The concept of subset is generalized to a {\em subobject classifier} in a topos.  To help build some intuition, consider how to define subsets without ``looking inside" a set. Essentially, a subset $S$ of some larger set $T$ can be viewed as a ``monic arrow", that is, an injective (or 1-1) function $f: S \hookrightarrow T$. 

\begin{definition} \citep{maclane:sheaves}
    An {\bf elementary topos} is a category ${\cal C}$ that has all (i) limits and colimits, (ii) has exponential objects, and (iii) a subobject classifier.
\end{definition} 
\begin{definition} 
    In a category ${\cal C}$ with finite limits, a {\bf subobject classifier} is a ${\cal C}$-object $\Omega$, and a ${\cal C}$-arrow ${\bf true}: {\bf 1} \rightarrow \Omega$, such that to every other monic arrow $S \hookrightarrow X$ in ${\cal C}$, there is a unique arrow $\phi$ that forms the following pullback square: 
\[\begin{tikzcd}
	S &&& {{\bf 1}} \\
	\\
	X &&& \Omega
	\arrow["m", tail, from=1-1, to=3-1]
	\arrow[from=1-1, to=1-4]
	\arrow["{{\bf true}}"{description}, tail, from=1-4, to=3-4]
	\arrow["{\phi}"{description}, dashed, from=3-1, to=3-4]
\end{tikzcd}\]  
\end{definition} 
This commutative diagram is a classic example of how to state a universal property: it enforces a condition that every monic arrow $m$ (i.e., every $1-1$ function) that maps a ``sub"-object $S$ to an object $X$ must be characterizable in terms of a ``pullback", a particular type of universal property that is a special type of a limit. In the special case of the category of sets, it is relatively easy to show that subobject classifiers are simply the characteristic (Boolean-valued) function $\phi$ that defines subsets. In general, as we show below, the subobject classifier $\Omega$ for causal models is not Boolean-valued, and requires using intuitionistic logic through a Heyting algebra.  This definition can be rephrased as saying that the subobject functor is representable. In other words, a subobject of an object $x$ in a category ${\cal C}$ is an equivalence class of monic arrows $m: S \hookrightarrow  x$. 

 \begin{theorem}\citep{mahadevan2025universalreinforcementlearningcoalgebras}
 \label{llmtopos}
     The category ${\cal C_Q}$  of value functions forms a topos. 
 \end{theorem}

\section{Long-Term Memory as a Topos of Coalgebras}

\label{ltm-topos} 

We now show how to model compositional structure in long-term memory by introducing {\em monoidal categories}, where objects can be multiplied, copied, and deleted. These operations are fundamental if the brain is to be capable of neurally realizing probabilistic inference.  First, we introduce how to model probabilistic reasoning using  symmetric monoidal categories, where every object has a comonoid defined on it \citep{Fritz_2020} that allow copying and deleting objects. We postulate that the brain must allow such operations to facilitate neural probabilistic reasoning. Then, we show that if the process of URL (and as a special case, RL) creates long-term optimal behavior by mapping sequential deliberative action sequences from short-term conscious memory into long-term asynchronous distributed memory, then it creates a topos of value functions (we showed this result in our previous paper \citep{mahadevan2025universalreinforcementlearningcoalgebras}). 

\subsection{Symmetric Monoidal Categories}

\label{smc}

\begin{figure}[h]
    \centering
    \includegraphics[width=0.75\linewidth]{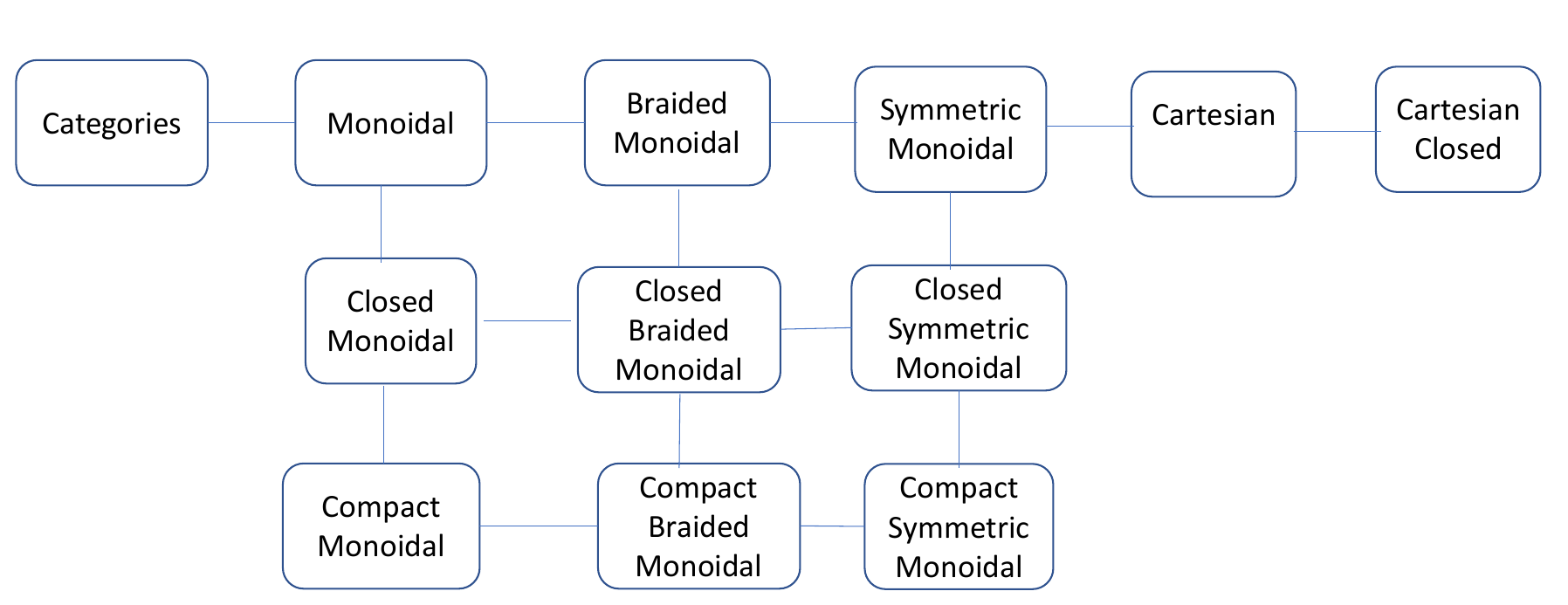}
    \caption{A fundamental requirement in modeling consciousness is that objects can be copied, multiplied, and deleted. We accomplish this categorically by using symmetric monoidal categories.}
    \label{monoidal-cats}
\end{figure}

 We briefly introduce the notion of symmetric monoidal categories (see Figure~\ref{monoidal-cats}), which are important for modeling consciousness as it enables us to copy, delete, and multiply objects. These type of operations are crucial if the brain is capable of realizing probabilistic computations. Good introductions are available in a number of textbooks \cite{maclane:71,richter2020categories}. 

\begin{definition}
    A {\bf monoidal category} is a category {\cal C} together with a functor $\otimes: {\cal C} \times {\cal C} \rightarrow {\cal C}$, an identity object $e$ of {\cal C} and natural isomorphisms $\alpha, \lambda, \rho$ defined as: 

    \begin{eqnarray*}
        \alpha_{C_1, C_2, C_3}: C_1 \otimes (C_2 \otimes C_3) & \cong & (C_1 \otimes C_2) \otimes C_2, \ \ \mbox{for all objects} \ \ C_1, C_2, C_3 \\
        \lambda_C: e \otimes C & \cong & C, \ \ \mbox{for all objects} \  \ C  \\
        \rho: C \otimes e & \cong & C, \ \  \mbox{for all objects} \ C
    \end{eqnarray*}
\end{definition}

The natural isomorphisms must satisfy coherence conditions called the ``pentagon" and ``triangle" diagrams \cite{maclane:71}. An important result shown in \cite{maclane:71} is that these coherence conditions guarantee that all well-formed diagrams must commute. 

\begin{definition}
    A {\bf symmetric monoidal category} is a monoidal category $({\cal C}, \otimes, e, \alpha, \lambda, \rho)$ together with a natural isomorphism 

    \begin{eqnarray*}
       \tau_{C_1, C_2}: C_1 \otimes C_2 \cong C_2 \otimes C_1, \ \ \mbox{for all objects} \ \ C_1, C_2
    \end{eqnarray*}
    where $\tau$ satisfies the additional conditions: for all objects $C_1, C_2$ $\tau_{C_2, C_1} \circ \tau_{C_1, C_2} \cong 1_{C_1 \otimes C_2}$, and for all objects $C$, $\rho_C = \lambda_C \circ \tau_{C, e}: C \otimes e \cong C$. 
\end{definition}

An additional hexagon axiom is required to ensure that the $\tau$ natural isomorphism is compatible with $\alpha$.  The $\tau$ operator is called a ``swap" in Markov categories \cite{Fritz_2020}. These isomorphisms are easier to visualize as string diagrams, as will be illustrated below in Section~\ref{mccat}. 

\subsection{Modeling Uncertainty using Markov Categories } 

\label{mccat}

If the brain is to be capable of modeling probabilistic computation, a fundamental result in categorical probability shows that it must have the capacity to multiply, copy, and delete objects \citep{Fritz_2020}. We introduce this framework of Markov categories, which provides the basis for reasoning about causality, decision making and uncertainty in our CF framework. Markov categories have been studied extensively as a unifying categorical model for causal inference, probability and statistics. They are symmetric monoidal categories, which we reviewed in Section~\ref{smc}, combined with a comonoidal structure on each object. Importantly, Markov categories are semi-Cartesian because they do not use uniform copying, but contain a Cartesian subcategory defined by deterministic morphisms.  We give a brief review of Markov categories, and significant additional details that are omitted can be found in \cite{Cho_2019,Fritz_2020,fritz:jmlr}.  For the sake of clarity, we follow the definitions in \cite{Fritz_2020}. 

\begin{definition}
	\label{markov_cat}
	A \emph{Markov category} ${\cal C}$ \citep{Fritz_2020} is a symmetric monoidal category in which every object $X \in {\cal C}$ is equipped with a commutative comonoid structure given by a comultiplication $\cop_X : X \to X \otimes X$ and a counit $\del_X : X \to I$, which satisfy commutative comonoid equations 
	as well as compatibility with the monoidal structure,
	and naturality of $\del$.
\end{definition}

We now discuss a subcategory of Cartesian categories within  Markov that involves uniform $\mbox{copy}_X$ and $\mbox{del}_X$ morphisms. One fundamental property of Markov categories is that they are {\em semi-Cartesian}, as the unit object is also a terminal object.  But, a subtlety arises in how these copy and delete operators are modeled, as we discuss below. 

\begin{definition}
    A symmetric monoidal category ${\cal C}$ is {\bf Cartesian} if the tensor product $\otimes$ is the categorical product. 
\end{definition}

If ${\cal C}$ and ${\cal D}$ are symmetric monoidal categories, then a functor $F: {\cal C} \rightarrow {\cal D}$ is monoidal if the tensor product is preserved up to coherent natural isomorphisms. $F$ is strictly monoidal if all the monoidal structures are preserved exactly, including $\otimes$, unit object $I$, symmetry, associative and unit natural isomorphisms. Denote the category of symmetric monoidal categories with strict functors as arrows as {\bf MON}. Let us review the basic definitions given by \citet{Heunen2019}, which will give some further clarity on the Cartesian structure in  Markov categories. 

\begin{definition}
 The subcategory of comonoids {\bf coMON} in the ambient category {\bf MON} of all symmetric monoidal categories is defined for any specific category {\cal C} as a collection of ``coalgebraic" objects $(X, \mbox{copy}_X, \mbox{del}_X)$, where $X$ is in {\cal C}, and arrows defined as comonoid homomorphisms  from $(X, \mbox{copy}_X, \mbox{del}_X)$ to  $(Y, \mbox{copy}_Y, \mbox{del}_Y)$ that act uniformly, in the sense that if $f: X \rightarrow Y$ is any morphism in {\cal C}, then: 

 \begin{eqnarray*}
     (f \otimes f) \circ \mbox{copy}_X = \mbox{copy}_Y \circ f \\
     \mbox{del}_Y \circ f = \mbox{del}_X 
 \end{eqnarray*}
\end{definition}

\citet{Heunen2019} define the process of ``uniform copying and deleting" in the category {\bf coMON}, which we now relate to Markov categories. A subtle difference worth emphasizing with Definition~\ref{markov_cat} is that in Markov categories, only $\mbox{del}_X$ is ``uniform", but not $\mbox{copy}_X$ in the sense defined by \citet{Heunen2019}. 

\begin{definition} \cite{Heunen2019}
    A symmetric monoidal category {\cal C} admits {\bf uniform deleting} if there is a natural transformation $e_X: X \xrightarrow{e_X} I$ for all objects in the  subcategory ${\cal C}_{\bf coMON}$ of comonoidal objects, where $e_I = \mbox{id}_I$. 
\end{definition}

Essentially, we require that if you process some object and then discard it, it's equivalent to discarding it without processing. 

\begin{theorem} \cite{Heunen2019}
A symmetric monoidal category {\cal C} has uniform deleting if and only if $I$ is terminal. 
\end{theorem}

This property holds for Markov categories, as noted in \cite{Fritz_2020}, and a simple diagram chasing proof is given in \cite{Heunen2019}. 

\begin{definition}\cite{Heunen2019}
    A symmetric monoidal category {\cal C} has {\bf uniform copying} if there is a natural transformation $\mbox{copy}_X: X \rightarrow X \otimes X$ such that $\mbox{del}_I = \rho^{-1}_I$.  
\end{definition}

We can now state an important result proved in \cite{Heunen2019} (Theorem 4.28), which relates to the more general results  shown earlier by \citet{fox}.  

\begin{theorem}\cite{fox,Heunen2019}
\label{heunen-fox}
    The following conditions are equivalent for a symmetric monoidal category {\cal C}. 

    \begin{itemize} 

    \item The category {\cal C} is {\bf Cartesian} with tensor products $\otimes$ given by the categorical product and the tensor unit is given by the terminal object. 

    \item The symmetric monoidal category {\cal C} has {\bf uniform copying and deleting}. 

    \end{itemize} 
\end{theorem}

As noted by \citet{Fritz_2020}, not all Markov categories are Cartesian, because their {\bf copy}$_X$ is not uniform, but only {\bf del}$_X$ is. For example, consider the  category {\bf FinStoch}, where a joint distribution is specified by the morphism $\psi: I \rightarrow X \otimes Y$. In this case, the marginal distributions can be formed as the composite morphisms
\begin{eqnarray*}
    I \xrightarrow{\psi} X \otimes  Y \xrightarrow{\mbox{del}_Y} X \\ 
    I \xrightarrow{\psi} X \otimes  Y \xrightarrow{\mbox{del}_X} Y \\ 
\end{eqnarray*}
But to require that in this case $\otimes$ is the categorical product implies that the marginal distributions defined as the above composites must be in bijection with the joint distribution. We showed previously \citep{cktheory} that Pearl's structural causal models (SCMs) are essentially Cartesian Markov categories. 

Recall from Theorem~\ref{heunen-fox} that  symmetric monoidal categories that are Cartesian admit uniform copying and deleting. 

\begin{definition}\cite{Fritz_2020}
A morphism $f: X \rightarrow Y$ in a  Markov category ${\cal C}$ is deterministic if it admits uniform copying. 
\end{definition}

We can now define a category of causal models in our CF framework as a Markov category where  all the morphisms between exogenous variables and endogenous variables are deterministic. 

\begin{definition}
    A structural causal model (SCM) \cite{pearl-book} can be defined as a restricted type of Markov category  ${\cal C_{SCM}}$ whose collection of objects $\mbox{Obj} = \{U, V \}$ is partitioned into a collection of exogenous objects $U$ and a collection of endogenous objects $V$, such that every morphism $f: X \rightarrow Y$ from an exogenous object $X \in U$ to an endogenous object $Y \in V$ is deterministic. 
\end{definition}

Observe that any product of exogenous objects $X_1  \otimes X_2 \ldots X_n$ is exogenous if each object $X_i$ is exogenous, and similarly, any product of endogenous objects is also endogenous if each object in the product is endogenous. Thus, every exogenous variable is defined with respect to some probability distribution $P$. 

\begin{definition}
    Given a cartesian Markov category ${\cal C_{SCM}}$ corresponding to a structural causal model, to every object corresponding to an exogenous variable $X$, there exists a morphism $\psi: I \rightarrow X$ that defines a distribution over $X$. 
\end{definition}

\subsection{The Topos of Coalgebras}

A fundamental result in topos theory states that for any given topos ${\cal E}$ that has a comonad $(G, \delta, \epsilon)$ defined on it itself induces another topos of coalgebras. This result shows why it is important to have the ``copy-delete" operation in Markov categories. Besides allowing for causal, probabilistic, and statistical reasoning, the comonoidal structure allows us to define logical reasoning via the internal language of the resulting topos. 

\begin{theorem}\citep{maclane:sheaves}
\label{lexthm}
    If $(G, \delta, \epsilon)$ is a {\em comonad} on a topos ${\cal E}$ for which the functor $G$ is {\em left exact}, then the category ${\cal E}_G$ of coalgebras for the comonad $(G, \delta, \epsilon)$ is itself a topos. 
\end{theorem}

A {\em left exact} functor is one that preserves all limits, whereas a right exact functor preserves colimits. 

We now turn to explaining what the internal language of a topos is, and introduce the MUMBLE internal language used in our CF framework.

\section{MUMBLE: Multi-modal Universal Mitchell-B\'enabou Language Embeddings}

\label{mumble}

In the CTM framework for modeling consciousness,  \citet{blum:pnas} introduced a internal language of thought \citep{fodor:loth} called ``Brainish". In our CF framework, the ``internal language of thought" \citep{fodor:loth} is defined by MUMBLE, which stands for Multi-modal Universal Mitchell-B/'enabou Language Embedding. This formal internal language is associated with every topos category.  we first need to formally define the mathematics of internal languages in a topos. We define formally what an internal local set theory is, and how it can be associated with an externally defined topos category. Our discussion draws from standard textbook treatments, including \cite{maclane:sheaves,Johnstone:topostheory}. We first define local set theories, and then define the Mitchell-B\'enabou internal language of a category and specify its Kripke-Joyal semantics. 

As Figure~\ref{fig:arch} illustrates, the flow of information into short-term deliberative conscious memory is intrinsically {\em multi-modal}, fusing together perception, motor control information, language, world knowledge, and many other components of the mind. For this plethora of processes to be able to internally ``talk" to each other in a common language, we need to define formally what such a language might look like. Our goal here is not neural plausibility, but mathematical clarity. What categorical structure admits such a broad set of inferential tools? We argue that it must be a topos structure, as it admits of a local set theory, and a logic with well-defined semantics. We explain the basic theory of an internal logic of a topos below. The material is standard, and can be found in many textbooks on topos theory \citep{maclane:sheaves,bell,Johnstone:topostheory}. 

\subsection{Local Set Theories} 

It is well-understood that properties of sets can be expressed as statements in first-order logic. For example, the following logical statement expresses a property of real numbers: 

\[ \forall x \ \exists y \ \ \ x < y \ \ \ x, y \in \mathbb{R} \]

namely that there does not exist a largest real number. In interpreting such logical statements, every variable $x, y, \ldots $ must be assigned a real number, and has to be interpreted as either ``free" or ``bound" by a quantifier. The above expression has no free variables. Each logical connective, such as $\leq$ must be also given an interpretation. The entire expression has to be assigned a ``truth value" in terms of whether it is true or false. In the development of the internal language associated with a topos, we will see that truth values are not binary, and can take on many possible values. In a presheaf category ${\bf Sets}^{{\cal C}^{op}}$, the subobject classifier $\Omega(C)$ of an object is defined as the partially ordered set of all subobjects, and its ``truth" value is not binary! It is possible to define a ``local set theory" that can be formulated without making any reference at all to sets, but merely as an axiomatic system over a set of abstract types, which will be interpreted in terms of the objects of a topos category below. We briefly sketch out the elements of a local set theory, and refer the reader to the details in \cite{bell}. 

A {\em local set theory} is defined as a language ${\cal L}$ specified by the following classes of symbols: 

\begin{enumerate}
    \item Symbols ${\bf 1} $ and $\Omega$ representing the {\em unity} type and {\em truth-value} type symbols. 

    \item A collection of symbols ${\bf A}, {\bf B}, {\bf C}, \ldots $ called {\em ground type symbols}. 

    \item A collection of symbols ${\bf f}, {\bf g}, {\bf h}, \ldots$ called {\em function} symbols. 
\end{enumerate}

We can use an inductive procedure to recursively construct {\bf type symbols} of ${\cal L}$ as follows: 

\begin{enumerate}
    \item  Symbols ${\bf 1} $ and $\Omega$ are type symbols.

    \item Any ground type symbol is a type symbol. 

    \item If ${\bf A}_1, \ldots, {\bf A}_n$ are type symbols, so is their product ${\bf A}_1 \times \ldots {\bf A}_n$, where for $n=0$, the type of $\prod_{i=1}^n {\bf A}_i$ is ${\bf 1}$. The product ${\bf A}_1 \times \ldots {\bf A}_n$ has the {\em product type} symbol. 

     \item If ${\bf A}$ is a type symbol, so is ${\bf PA}$. The type ${\bf PA}$ is called the {\em power} type. \footnote{Note that in a topos, these will be interpreted as {\em power objects}, generalizing the concept of power sets.}
\end{enumerate}

For each type symbol ${\bf A}$, the language ${\cal L}$ contains a set of {\em variables} $x_{\bf A}, y_{\bf A}, z_{\bf A}, \ldots$. In addition, ${\cal L}$ contains the distinguished ${\bf *}$ symbol. Each function symbol in ${\cal L}$  is assigned a {\em signature} of the form ${\bf A} \rightarrow {\bf B}$. \footnote{In a topos, these will correspond to arrows of the category.} We can define the {\em terms} of the local set theory language ${\cal L}$ recursively as follows: 

\begin{itemize}
    \item ${\bf *}$ is a term of type ${\bf 1}$. 

    \item for each type symbol ${\bf A}$, variables $x_{\bf A}, y_{\bf A}, \ldots$ are terms of type ${\bf A}$. 

    \item if ${\bf f}$ is a function symbol with signature ${\bf A} \rightarrow {\bf B}$, and $\tau$ is a term of type ${\bf A}$, then ${\bf f}(\tau)$ is a term of type ${\bf B}$. 

    \item If $\tau_1, \ldots, \tau_n$ are terms of types ${\bf A}_1, \ldots, {\bf A}_n$, then $\langle \tau_1, \ldots \tau_n \rangle$ is a term of type ${\bf A}_1 \times \ldots {\bf A}_n$, where if $n=0$, then $\langle \tau_1, \ldots \tau_n \rangle$ is of type ${\bf *}$. 

    \item If $\tau$ is a term of type ${\bf A}_1 \times {\bf A}_n$, then for $1 \leq i \leq n$, $(\tau)_i$ is a term of type ${\bf A}_i$. 

    \item if $\alpha$ is a term of type $\Omega$, and $x_{\bf A}$ is a variable of type ${\bf A}$, then $\{x_{\bf A} : \alpha \}$ is a term of type ${\bf PA}$.  

    \item if $\sigma, \tau$ are terms of the same type, $\sigma = \tau$ is a term of type $\Omega$. 

    \item if $\sigma, \tau$ are terms of the types ${\bf A}, {\bf PA}$, respectively, then $\sigma \in \tau$ is a term of type ${\bf \Omega}$. 
\end{itemize}

A term of type ${\bf \Omega}$ is called a {\em formula}. The language ${\cal L}$ does not yet have defined any logical operations, because in a typed language, logical operations can be defined in terms of the types, as illustrated below. 

\begin{itemize}
    \item $\alpha \Leftrightarrow \beta$ is interpreted as $\alpha = \beta$. 

    \item {\bf true} is interpreted as ${\bf *} = {\bf *}$. 

    \item $\alpha \wedge \beta$ is interpreted as $\langle \alpha, \beta \rangle = \langle {\bf true}, {\bf false} \rangle$. 

    \item $\alpha \Rightarrow \beta$ is interpreted as $(\alpha \wedge \beta) \Leftrightarrow \alpha$

    \item $\forall x \ \alpha$ is interpreted as $\{x : \alpha\} = \{x : {\bf true} \}$

    \item ${\bf false}$ is interpreted as $\forall \omega \ \omega$. 

    \item $\neg \alpha$ is interpreted as $\alpha \Rightarrow {\bf false}$.

    \item $\alpha \vee \beta$ is interpreted as $\forall \omega \ [(\alpha \Rightarrow \omega \wedge \beta \Rightarrow \omega) \Rightarrow \omega]$

    \item $\exists x \ \alpha$ is interpreted as $\forall \omega [ \forall x (\alpha \Rightarrow \omega) \Rightarrow \omega ]$

\end{itemize}

Finally, we have to specify the inference rules, which are given in the form of {\em sequents}. We will just sketch out a few, and the rest can be seen in \cite{bell}. A sequent is a formula 

\[ \Gamma: \alpha\]

where $\alpha$ is a formula, and $\Gamma$ is a possibly empty finite set of formulae. The basic axioms include $\alpha: \alpha$ (tautology), $:x_1 = {\bf *}$ (unity), a rule for forming projections of products, a rule for equality, and another for comprehension. Finally, the inference rules are given in the form: 

\begin{itemize}
    \item {\em Thinning:}
    \[
  \begin{prooftree}
    \hypo{\Gamma : \alpha}
    \infer1{\beta, \Gamma: \alpha}
  \end{prooftree}
\]
\item {\em Cut}: 

\[
  \begin{prooftree}
    \hypo{\Gamma : \alpha, \  \ \alpha, \Gamma: \beta}
    \infer1{\Gamma: \beta}
  \end{prooftree}
\]

\item {\em Equivalence}: 

\[
  \begin{prooftree}
    \hypo{\alpha, \Gamma : \beta \ \ \beta, \Gamma: \alpha}
    \infer1{\Gamma: \alpha \Leftrightarrow \beta}
  \end{prooftree} 
\]

\end{itemize}

A full list of inference rules with examples of proofs is given in \cite{bell}. Now that we have the elements of a local set theory defined as shown above, we need to connect its definitions with that of a topos. That is the topic of the next section.

\subsection{Mitchell-B\'enabou Language of a Topos}
\label{mbl}

We now define the central theoretical core of internal languages of thought in terms of the objects and arrows of a topos category, which are commonly referred to as the Mitchell-B\'enabou languages (MBL) \cite{mitchell:topoi}.  As with the abstract local set theory defined in the previous section, we have to define the types (which will be the objects of a topos), the functions and terms, and give definition of universal and existential quantifiers. We postpone the discussion of the interpretation of this language to the next section. 

Given a topos category ${\cal C}$, we define the types of MBL as the objects of ${\cal C}$. Note that for a presheaf category $\hat{{\cal C}} = {\bf Sets}^{{\cal C}^{op}}$, the types will correspond to the functor objects given by the Yoneda embedding $\yo(x) = {\cal C}(-, x)$ (contravariantly) or $\yo(x) = {\cal C}(x, -)$. Instantiating this process for an LLM category \cite{bradley:enriched-yoneda-llms}, note that for a given text fragment, such as 

\[ x = \mbox{I drove}\]

its continuation $y$ could mean many phrases including examples such as 

\[ y = \mbox{to work} \]

For each type $C$ (defined as an object of the topos category ${\cal C}$), like for a local set theory, we assume the existence of variables $x_C, y_C, \ldots$, where each such variable has as its interpretation the identity arrow ${\bf 1}: C \rightarrow C$. Just like for local set theories, we can construct product objects, such as $A \times B \times C$, where terms like $\sigma$ that define arrows are given the interpretation 

\[ \sigma: A \times B \times C \rightarrow D \]

We can inductively define the terms and their interpretations in a topos category as follows (see \cite{maclane:sheaves} for additional details): 

\begin{itemize}
    \item Each variable $x_C$ of type $C$ is a term of type $C$, and its interpretation is the identity $x_C = {\bf 1}: C \rightarrow C$. 

\item Terms $\sigma$ and $\tau$ of types $C$ and $D$ that are interpreted as $\sigma: A \rightarrow C$ and $\tau: B \rightarrow D$ can be combined to yield a term $\langle \sigma, \tau \rangle$ of type $C \times D$, whose joint interpretation is given as 

\[ \langle \sigma p, \tau q \rangle: X \rightarrow C \times D\]

where $X$ has the required projections $p: X \rightarrow A$  and $q: X \rightarrow B$. 

\item Terms $\sigma: A \rightarrow B$ and $\tau: C \rightarrow B$ of the same type $B$ yield a term $\sigma = \tau$ of type $\Omega$, interpreted as 

\[ (\sigma = \tau): W \xrightarrow[]{\langle \sigma p, \tau q \rangle} B \times B \xrightarrow[]{\delta_B} \Omega \]

where $\delta_B$ is the characteristic map of the diagonal functor $\Delta B \rightarrow B \times B$. In the AGI modality for causal inference,  these diagonal maps will correspond to the ``copy" procedure in a topos category of presheaves over Markov categories \cite{Fritz_2020}. 

\item Arrows $f: A \rightarrow B$ and a term $\sigma: C \rightarrow A$ of type $A$ can be combined to yield a term $f \circ \sigma$ of type $B$, whose interpretation is naturally a composite arrow: 

\[ f \circ \sigma: C \xrightarrow[]{\sigma} A \xrightarrow[]{f} B\]

\item For exponential objects, terms $\theta: A \rightarrow B^C$ and $\sigma: D \rightarrow C$ of types $B^C$ and $C$, respectively, combine to give an ``evaluation" map of type $B$, defined as 

\[ \theta (\sigma): W \rightarrow B^C \times C \xrightarrow[]{e} B \]

where $e$ is the evaluation map, and $W$ defines a map $\langle \theta p, \sigma q \rangle$, where once again $p: W \rightarrow A$ and $q: W \rightarrow D$ are projection maps. 

\item Terms $\sigma: A \rightarrow B$ and $\tau: D \rightarrow \Omega^B$ combine to yield a term $\sigma \in \tau$ of type $\Omega$, with the following interpretation: 

\[ \sigma \in \tau: W \xrightarrow[]{\langle \sigma p, \tau q \rangle} B \times \Omega^B \xrightarrow[]{e} \Omega \]

\item Finally, we can define local functions as $\lambda$ objects, such as 

\[ \lambda x_C \sigma: A \rightarrow B^C \]

where $x_C$ is a variable of type $C$ and $\sigma: C \times A \rightarrow B$. 
\end{itemize}

Once again, we can combine terms $\alpha, \beta$ etc. of type $\Omega$ using logical connectives $\wedge, \vee, \Rightarrow, \neg$, as well as quantifiers, to get composite terms, where each of the logical connectives is now defined over the subobject classifier $\Omega$, giving us

\begin{itemize}
    \item $\wedge: \Omega \times \Omega \rightarrow \Omega$ is interpreted as the {\em meet} operation in the partially ordered set of subobjects (given by the Heyting algebra). 

    \item $\vee: \Omega \times \Omega \rightarrow \Omega$ is interpreted as the {\em join} operation in the partially ordered set of subobjects (given by the Heyting algebra). 

    \item $\Rightarrow: \Omega \times \Omega \rightarrow \Omega$ is interpreted as an adjoint functor, as defined previously for a Heyting algebra. 
    
\end{itemize}

We can combine these logical connectives with the term interpretation as arrows as defined earlier in a fairly straightforward way, as described in \cite{maclane:sheaves}. We now turn to the Kripe-Joyal semantics of this language. 

\subsection{Kripke-Joyal Semantics}
\label{kj}

 Let ${\cal C}$ be a topos, and let it possess a Mitchell-B\'enabou language as defined above. How do we define a suitable model for this language? In this section, we define the Kripke-Joyal semantics that provides an interpretation of the Mitchell-B\'enabou language described in the previous section. A more detailed overview  of this topic is given in \cite{maclane:sheaves}.

 For the category ${\cal C}$, and for any object $X$ in ${\cal C}$, define a {\em generalized element} as simply a morphism $\alpha: U \rightarrow X$. We want to specify the semantics of how $U$ supports any formula $\phi(\alpha)$, denoted by $U \Vdash \phi(\alpha)$. We declare that this ``forcing" relationship holds if and only if $\alpha$ factors through $\{x | \phi(x) \}$, where $x$ is a variable of type $X$ (recall that objects $X$ of a topos form its types), as shown in the following commutative diagram. 

 \begin{center}
\begin{tikzcd}
	&& {\{x | \phi(x) \}} && {{\bf 1}} \\
	\\
	U && X && {{\bf \Omega} }
	\arrow[from=1-3, to=1-5]
	\arrow[tail, from=1-3, to=3-3]
	\arrow["{{\bf True}}", from=1-5, to=3-5]
	\arrow[dashed, from=3-1, to=1-3]
	\arrow["\alpha"', from=3-1, to=3-3]
	\arrow["{\phi(x)}", from=3-3, to=3-5]
\end{tikzcd}
 \end{center}

 Building on this definition, if $\alpha, \beta: U \rightarrow X$ are parallel arrows, we can give semantics to the formula $\alpha = \beta$ by the following statement: 

 \[ U \xrightarrow[]{\langle \alpha, \beta \rangle} X \times X \xrightarrow[]{\delta_X} \Omega\]

following the definitions in the previous section for the composite $\langle \alpha, \beta \rangle$ and $\delta_X$ in MBL. 

We can extend the previous commutative diagram to show that $U \Vdash \alpha = \beta$ holds if and only if $\langle \alpha, \beta \rangle$ factors through the diagonal map $\Delta$: 

\begin{center}
\begin{tikzcd}
	&& X && {{\bf 1}} \\
	\\
	U && {X \times X} && {{\bf \Omega} }
	\arrow[from=1-3, to=1-5]
	\arrow["\Delta", tail, from=1-3, to=3-3]
	\arrow["{{\bf True}}", from=1-5, to=3-5]
	\arrow[dashed, from=3-1, to=1-3]
	\arrow["{\langle \alpha, \beta \rangle}"', from=3-1, to=3-3]
	\arrow["{\delta_x}", from=3-3, to=3-5]
\end
{tikzcd}
\end{center}

Many additional properties can be derived (see \cite{maclane:sheaves}), including the following useful ones. 

\begin{itemize}
    \item {\bf Monotonicity:} If $U \Vdash \phi(x)$, then we can pullback the interpretation through any arrow $f: U' \rightarrow U$ in a topos ${\cal C}$ to obtain $U' \Vdash \phi(\alpha \circ f)$. 
    \begin{center}
\begin{tikzcd}
	&&&& {\{x | \phi(x) \}} && {{\bf 1}} \\
	\\
	{U'} && U && X && {{\bf \Omega} }
	\arrow[from=1-5, to=1-7]
	\arrow[tail, from=1-5, to=3-5]
	\arrow["{{\bf True}}", from=1-7, to=3-7]
	\arrow[dashed, from=3-1, to=1-5]
	\arrow["f", from=3-1, to=3-3]
	\arrow[dashed, from=3-3, to=1-5]
	\arrow["\alpha"', from=3-3, to=3-5]
	\arrow["{\phi(x)}", from=3-5, to=3-7]
\end{tikzcd}
    \end{center}

    \item {\bf Local character:} Analogously, if $f: U' \rightarrow U$ is an epic arrow, then from $U' \Vdash \phi(\alpha \circ f)$, we can conclude $U \Vdash \phi(x)$. 
\end{itemize}

We can summarize the main results of Kripke-Joyal semantics using the following theorem. These give precise semantics for the standard logical connectives, as well as universal and existential quantification in terms of the arrows of a topos category ${\cal C}$. We can specialize these broad results to specific AGI categories in the subsequent sections. 

\begin{theorem}\cite{maclane:sheaves}
    If $\alpha: U \rightarrow X$ is a generalized element of $X$, and $\phi(x)$ and $\psi(x)$ are formulas with a free variable $x$ of type $X$, we can conclude that
    \begin{enumerate}
        \item $U \Vdash \phi(\alpha) \wedge \psi(\alpha)$ holds if $U \Vdash \phi(\alpha)$ and $U \Vdash \psi(\alpha)$. 
        \item $U \Vdash \phi(x) \vee \psi(x)$ holds if there are morphisms $p: V \rightarrow U$ and $q: W \rightarrow U$ such that $p + q: V + W \rightarrow U$ is an epic arrow, and $V \Vdash \phi(\alpha p)$ and $W \Vdash \phi(\alpha q)$. 
        \item $U \Vdash \phi(\alpha) \Rightarrow \psi(\alpha)$ if it holds that for any morphism $p: V \rightarrow U$, where $V \Vdash \phi(\alpha p)$, the assertion $V \Vdash \phi(\alpha p)$  also holds. 

        \item $U \Vdash \neg \phi(\alpha)$ holds if whenever the morphism $p: U \rightarrow V$ satisfies the property $V \Vdash \phi(\alpha p)$, then $V \cong {\bf 0}$. 

        \item $U \Vdash \exists \phi(x,y)$ holds if there exists an epic arrow $p: V \rightarrow U$ and generalized elements $\beta: V \rightarrow Y$ such that $V \Vdash \phi(\alpha p, \beta)$. 

        \item $U \Vdash \forall y \phi(x, y)$ holds if for every object $V$, and every arrow $p: V \rightarrow U$, and every generalized element $\beta: V \rightarrow Y$, it holds that $V \Vdash \phi(\alpha p, \beta)$. 
    \end{enumerate}
\end{theorem}

To understand the significance of this theorem, note that we can now use it to provide rigorous semantics for how conscious and unconscious processes modeled in the category of coalgebras can ``talk" with one another in an internal topos language. 

\subsection{Kripke-Joyal Semantics for Sheaves}

Define ${\tt Sh}({\cal C, J})$ be a topos of sheaves with a specified Grothendieck topology ${\cal J}$, defined by the following diagram: 

\[ {\cal C} \xrightarrow[]{\yo} {\cal P(C)} \xrightarrow[]{a} {\tt Sh}({\cal C,J}) \cong {\cal C}\]

where we know that the Yoneda embedding $\yo$ creates a full and faithful copy of the original category ${\cal C}$. Let us define the semantics for a sheaf element $\alpha \in X(C)$, where $ X(C) = {\tt Sh}({\cal C}, J)({\cal C}(-, C), X))$. Since we know that $\{x | \phi(x) \}$ is a subsheaf, and given an arrow $f: D \rightarrow C$ of ${\cal C}$, and $\alpha \in X(C)$, then if $\alpha$ is one of the elements that satisfies the property that $\{x | \phi(x) \}$, the monotonicity property stated above implies that $\alpha \circ f \in \{ x | \phi(x) \}(D) \subseteq X(D)$. Also, the local character condition stated above implies that if $\{f_i: C_i \rightarrow C \}$ is a cover in the Grothendieck topology ${\cal J}$ such that $C_i | \Vdash \phi(\alpha \circ f_i)$ for all $i$, then $C \Vdash \phi(\alpha)$. 

With these assumptions, we can restate the Kripke-Joyal semantics for the topos category of sheaves as follows: 

\begin{enumerate}
    \item $C \Vdash \phi(\alpha) \wedge \psi(\alpha)$ if it holds that $C \Vdash \phi(\alpha)$ and $C \Vdash \psi(\alpha)$. 
    \item $C \Vdash \phi(\alpha) \vee \psi(\alpha)$ if there is a covering $\{ f_i: C_i \rightarrow C \}$ such that for each $i$, either $C_i \Vdash \phi(\alpha)$ or $C_i \Vdash \psi(\alpha)$. 
    \item $C \Vdash \phi(\alpha) \rightarrow \psi(\alpha)$ if for all $f: D \rightarrow C$, and $D \Vdash \phi(\alpha \circ f)$, it holds that $D \Vdash \psi(\alpha \circ f)$. 
    \item $C \Vdash \neg \phi(\alpha)$ holds if for all arrows $f: D \rightarrow C$ in ${\cal C}$, if $D \Vdash \phi(\alpha \circ f)$ holds, then the empty family is a cover of $D$. 
    \item $C \Vdash \exists y \ \phi(x, y)$ holds if there is a covering $\{ f_i: C_i \rightarrow C \}$ and elements $\beta_i \in Y(C_i)$ such that $C_i \Vdash \phi(\alpha \circ f_i, \beta_i)$ holds for each $i$. 
    \item Finally, for universal quantification, $C \Vdash \forall y \ \phi(x, y)$ holds if for all arrows $f: D \rightarrow C$ in the category ${\cal C}$, and all $\beta \in Y(D)$, it holds that $D \Vdash \phi(\alpha \circ f, \beta)$. 
\end{enumerate}

Summarizing this somewhat abstract section, we began by defining a local set theory of types, within which we were able to state the language ${\cal L}$ and its inference rules. These abstractly characterize what a ``set-like" category should behave as. Subsequently, we showed that the Mitchell-B\'enabou language for a topos is precisely of the form of a local set theory, formalizing the precise way in which a topos is like a category of sets. Finally, we specified the Kripke-Joyal semantics for the Mitchell-B\'enabou internal language of a topos, and we also showed that for the specific case of sheaves constructed with the $\yo$ Yoneda embedding, what the resulting semantics looked like. 

We now have a precise semantics for MUMBLE'ing, namely the internal language of a topos, which can now serve as a ``language of thought" for our CF framework. 

\section{Mapping Conscious to Unconscious Processes using Universal RL} 
\label{udm} 

Now we turn to give a more detailed account of the process by which the slow deliberative trial-and-error nature of conscious short-term memory can be compiled into fast highly parallel, asynchronous and distributed long-term memory. Our approach builds on mathematical models of asynchronous parallel distributed computation \citep{bertsekas:pdc} as well as asynchronous decentralized decision making \citep{witsenhausen:1975}. 

Modeling consciousness requires modeling asynchronous distributed computation over many unconscious processes. There is a long history in computer science of modeling parallel distributed computation, and we build on the insights developed in two particular frameworks, one proposed by \citet{bertsekas:pdc}, and the other in the {\em intrinsic model} of \citet{witsenhausen:1975}. Our consciousness framework generalizes these to coalgebras, and we briefly describe the ideas here, before instantiating them in the context of modeling consciousness. Unlike the CTM framework, we do not posit any global clock that ticks at regular intervals and coordinates processor activity. The computation of various unconscious processors is completely asychronous, distributed, and parallel. We explain how it is possible to achieve this using two formal frameworks that have addressed this challenge in past work. 

The essence of asynchronous distributed computation is to manage the ensemble of ``processors" that are collectively computing some quantity. In the case of consciousness, different parts of the brain that are engaged in unconscious activity are monitoring many systems, such as visual fields, hearing, touch, motor sensations, language and so on. Each module works in parallel, but must compete with the others to post information into short-term memory.  We begin by introducing two simple and elegant approaches that address very specific instantiations of the more general problem. 

\subsection{Asynchronous Distributed Minimization} 

\citet{bertsekas:pdc} studied a generic asynchronous distributed computation of solving a fixed point equation 

\[ F(x^*) = x^*, \ \ \ x^* \in \mathbb{R}^n\]

where the mapping $F$ is comprised of a set of component mappings $f_i$, which admit asynchronous parallel distributed computation.  In the language of coalgebras, this fixed point equation will be described as finding a {\em final coalgebra}, a problem that has been solved in great generality \citep{Aczel1988-ACZNS}. The space of solutions that $x^*$ lies in is assumed to be the Cartesian product 

\[ X = X_1 \times X_2 \ldots X_n \]

and the solutions are vectors of the form 

\[ x = (x_1, \ldots, x_n) \]

where the component functions $f_i$ assemble together as 

\[ F(x) = (f_1(x), \ldots, f_n(x)), \ \ \ \forall \ x \in X \]

and the fixed point of $F$ is computed using an asynchronous distributed version of the iterative method 

\[ x_i = f_i(x) ,  \ \ \ i=1, \ldots, n \]

Algorithm~\ref{adm} describes a formulation of asynchronous distributed minimization of a decomposable function, partly rephrased in terms of the coalgebraic language we will use in the URL framework. 

\begin{algorithm}
\caption{Asynchronous Distributed Minimization \citep{bertsekas:pdc}} 
\label{adm}

{\bf Input:} Some function $F: X \rightarrow X$, where $X = \mathbb{R}^n$ or some (generalized) metric space decomposable as $X = X_1 \times X_2 \times \cdots, X_n$, and $F = \left( f_1(x), f_2(x), \ldots, f_n(x) \right)$, for all $x \in X$. 

{\bf Output:} A fixed point of $F$, namely an element $x^* \in X$ such that $F(x^*) = x^*$, which can be expressed also as $x_i^* = f_i(x^*)$, for all $i = 1, \ldots n$. This problem can be stated in the coalgebraic language of URL as finding a final coalgebra of a decomposable coalgebra. 

\begin{algorithmic}[1]
\REPEAT

\STATE At each time step $t \in T$, where $T = \{0, 1, \ldots, \}$, update some component $x_i$ of $x$ , using an asynchronous distributed version of the coalgebraic iteration: 

\[ x_i \rightarrow f_i(x) \]
\label{step1}

\STATE Each update  of $x_i$ is done in parallel by some ``processor" that may not have access to the latest values of all components of $x$, but we can assume that 

\[ x_i(t+1) = f_i(x1(\tau^1_i(t)), \ldots x_n(\tau^i_n(t))),  \ \ \forall t \in T^i\]

\STATE where $T^i$ is the set of time points where $x_i$ is updated, and $0 \leq \tau^i_j(t)  \leq t$, and at all times $t \notin T^i$, we assume that 

\[ x_i(t+1) = x_i(t) \]

\IF{the fixed point of $F$ is not reached}

\STATE Set $t = t+1$, and return to Step~\ref{step1}. 

\ELSE 

\STATE Set {\bf done} $\leftarrow$ {\bf true}. 

\ENDIF 

\UNTIL{{\bf done}}. 

\STATE  Return the fixed point $x^*$ of $F$. 

\end{algorithmic}
\end{algorithm} 

\subsection{Witsenhausen's Intrinsic Model} 

To manage the multiagent competition that occurs between unconscious processes modeled by the topos of coalgebras, we build on the intrinsic model.  \citet{witsenhausen:1975} introduced an elegant model of asynchronous distributed multiagent decision making in his {\em intrinsic model}, which we generalized to the categorical setting previously \citep{sm:udm}. We briefly summarize Witsenhausen's framework as defined in our Universal Decision Model framework, and explain its relevance to modeling decision making in consciousness. For a more detailed introduction to the Witsenhausen framework, we refer to the book on multistage optimization \citep{carpentier2015stochastic}. 

We briefly explain our previous work on a categorial generalization of Witsenhausen's framework, which we termed the Universal Decision Model (UDM)\cite{sm:udm}. In the UDM category ${\cal C}_{\mbox{UDM}}$, as in any category, we are given a collection of {\em decision objects} ${\cal D}$, and a set of morphisms ${\cal M}_{\mbox{UDM}}$ between UDM objects, where $f: c \rightarrow d$ is a morphism that maps from UDM object $c$ to $d$. A morphism need not exist between every pair of UDM objects. In this paper, we restrict ourselves to {\em locally small} UDM categories, meaning that exists only a set's worth of morphisms between any pair of UDM objects. 
 
 \begin{definition}  
 \label{udm-defn}
 A Universal Decision Model (UDM)  is defined as a category ${\cal C}_{\mbox{UDM}}$, where each decision object is represented as a  tuple $\langle (A, (\Omega, {\cal B}, P), U_\alpha, {\cal F}_\alpha, {\cal I}_\alpha)_{\alpha \in A} \rangle$, where $A$ in URL represent coalgebras,  $(\Omega, {\cal B}, P)$ is a probability space representing the inherent stochastic state of nature due to randomness,  $U_\alpha$ is a measurable space from which a decision $u \in U_\alpha$ is chosen by decision object $\alpha$. Each element's policy in a  decision object is any function $\pi_\alpha: \prod_\beta U_\beta \rightarrow U_\alpha$ that is measurable from its information field ${\cal I}_\alpha$, a subfield of  the overall product space $(\prod_\alpha U_\alpha, \prod_\alpha {\cal F}_\alpha)$, to the $\sigma$-algebra ${\cal F}_\alpha$. The policy of decision object $\alpha$ can be any function $\pi_\alpha: \prod_\beta U_\beta \rightarrow U_\alpha$. 
\end{definition} 

\begin{definition}  
The {\bf  information field} of an element $\alpha \in A$ in a decision object $c$ in UDM category ${\cal C}_{\mbox{UDM}}$ is denoted as ${\cal I}_\alpha \subset {\cal F}_A(A)$ characterizes the information available to decision object $\alpha$ for choosing a decision $u \in U_\alpha$. 
\end{definition}  

To ground this definition out in terms of the stochastic approximation theory of $Q$-learning, an information field precisely delineates what information is available to each of the parallel asynchronous distributed processors that are updating the $Q$-function. 
The information field structure yields a surprisingly rich topological space that has many important consequences for how to organize the  decision makers in a complex organization into subsystems. An element $\alpha$ in a decision object requires information from other elements or subsystems in the network. To formalize this notion, we use product decision fields and product $\sigma$-algebras, with their canonical projections.

\begin{definition}  
  Given a subset of nodes $B \subset A$, let $H_B = \Omega \times \prod_{\alpha \in B} U_\alpha$ be the {\bf product space of decisions} of nodes in the subset $B$, where the {\bf product $\sigma$-algebra} is ${\cal B} \times \prod_{\alpha \in B} {\cal F}_\alpha = {\cal F}_B(B)$. It is common to also denote the product $\sigma$-algebra by the notation $\otimes_{\alpha \in A} {\cal F}_\alpha$.  If $C \subset B$, then the {\bf induced $\sigma$-algebra} ${\cal F}_B(C)$ is a subfield of ${\cal F}_B(B)$, which can also be viewed as the inverse image of ${\cal F}_C(C)$ under the canonical projection of $H_B$ onto $H_C$. \footnote{Note that for any cartesian product of sets $\prod_i X_i$, we are always able to uniquely define  a projection map into any component set $X_i$, which is a special case of the product universal property in a category.} 
 \end{definition}

\subsection{Universal Reinforcement Learning:  Asynchronous Distributed Computation in Coalgebras}

As Figure~\ref{fig:arch} shows, our framework for consciousness models unconscious processes by coalgebras, which collectively must compete with each other to post their data in short-term memory. In the CTM, this process is modeled by a binary tree. In our framework, this tree data structure is generalized to a functor diagram, over which we apply our recent Universal Reinforcement Learning (URL) framework to model the competition. In this section, we briefly review URL, and refer the reader to the longer paper for additional details \citep{mahadevan2025universalreinforcementlearningcoalgebras}.

We can briefly summarize the main points of this paper by rephrasing Algorithm~\ref{adm} in the general coalgebraic setting as shown below. The problem of minimization of a vector function $F: X \rightarrow X$ is generalized to finding a final coalgebra with a specified $F$-dynamics, where $F$ is some functor in a symmetric monoidal category ${\cal C}$ with tensor product $\otimes$. 

\begin{algorithm}
\caption{Universal Reinforcement Learning over Coalgebras}
\label{urlalgm}

{\bf Input:} Given a functor diagram $F: {\cal J} \rightarrow {\cal C}$, where ${\cal C}$ is a category of coalgebras $\alpha_f = X \rightarrow f(X)$ that is (co)complete, meaning that it has categorical (co)products, (co)equalizers, and more generally has all finite (co)limits. This property ensures that any diagram $F$ is ``solvable" by finding its limit $\lim_F \in {\cal C}$. To concretize this abstraction, note that Algorithm~\ref{adm} defined a product function $F$, which can be seen as the limit of the diagram ${\cal J} = \bullet \bullet \cdots \bullet$. For example, in Section~\ref{topos-url}, we showed the category ${\cal C}_Q$ of action-value functions defines a topos, so that the coalgebras over ${\cal C}_Q$ will in addition to having (co)limits, also have a subobject classifier and exponential objects as part of their diagrammatic vocabulary. 

{\bf Output:} A final coalgebra $X^* \rightarrow F(X^*)$  such that $X^* \cong F(X^*)$ (where $\cong$ is the isomorphism in ${\cal C}$. 

\begin{algorithmic}[1]
\REPEAT

\STATE At each time step $t \in T$, where $T = \{0, 1, \ldots, \}$, update some component coalgebra $\alpha_{f_c} = X \rightarrow f_c(X)$ using an asynchronous distributed coalgebraic iteration, where $c$ is an object in ${\cal J}$.

\[ X_c \rightarrow f_c(X) \]
\label{step1url}

\STATE Each update  of $X$ is done in parallel by some ``processor" whose ``information field" is measurable (see Section~\ref{udm}). 

\[ X_c(t+1) = f(X_1(\tau^1_i(t)), \ldots X_{c_f}(\tau^i_n(t))),  \ \ \forall t \in T^i\]

where $c_f$ is the number of elements in the information field of element $c$. 

\STATE where $T^i$ is the set of time points where $X_f$ is updated, and $0 \leq \tau^i_j(t)  \leq t$, and at all times $t \notin T^i$, we assume that 

\[ X_f(t+1) = X_f(t) \]

\IF{the final coalgebra  $X^* \rightarrow F(X^*)$ is not found}

\STATE Set $t = t+1$, and return to Step~\ref{step1url}. 

\ELSE 

\STATE Set {\bf done} $\leftarrow$ {\bf true}. 

\ENDIF 

\UNTIL{{\bf done}}. 

\STATE  Return the final coalgebra $F(X^*) \cong X^*$ of $F$. 

\end{algorithmic}
\end{algorithm} 

\section{Mapping Unconscious to Conscious Memory as a Network Economy}
\label{network-economy-as-consciousness}

In the final section, we explore modeling the process of multiagent decision making among the large number of unconscious processes, modeled as coalgebras, as a {\em network economy} \citep{nagurney:vibook}. The fundamental problem is that short-term memory is a constrained resource, and we believe it is natural to explore applying the principle of network economics to this problem. In the CTM model, this competition is modeled using a binary tree, but we prefer to explore a broader framework that is amenable to asynchronous decentralized competition, following the discussion in the Section~\ref{udm} above. Broadly, a network economy consists of a group of autonomous agents that share a network, which are divided into producer agents, transporter agents, and consumer agents. To draw the parallel to modeling consciousness, the producer agents are the unconscious processes. The transporter agents are in charge of the problem of transporting (i.e., in the brain, the neural pathways leading to the area where conscious short-term memory resides) information from the unconscious long-term memory into short-term memory. Finally, the consumer agents are locations in short-term memory that can be seen as ``bidding" for different combinations of producer and transporter agents. This network economy framework can be seen as a type of evolutionary framework for modeling consciousness, broadly related to the work described by \citet{edelman:consciousness}.

\subsection{A Network Economic Model of Consciousness} 

Let us consider modeling consciousness as a collection of {\em producer} agents that want to post information in short-term memory from their unconscious processing (e.g., recalling some information from long-term memory),  a set of {\em transporter} agents that manage the neural pathways from long-term memory into short-term memory whose task is to transport the information generated by the producer agents, and finally a set of {\em consumer} agents corresponding to locations in short-term memory that must choose products from some combination of producer and transporter agents. The broad idea here is that information filters into short-term memory from long-term memory through a competitive bidding process. The mathematics of network economics involves {\em variational inequalities} (VIs) \citep{nagurney:vibook}, and we will explore an asynchronous decentralized framework for solving VIs using the type of approach described previously in Section~\ref{udm}. 

\begin{figure}[h]
\begin{center} \hfill
\begin{minipage}{0.95\textwidth}
\includegraphics[scale=0.6]{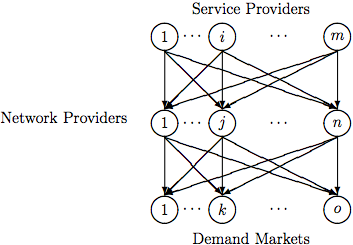}
\end{minipage} 
\end{center}
\caption{Modeling consciousness as a network economy \citep{nagurney:vibook}.The top tier represent producer agents from long-term unconscious memory who want to post their data into short-term conscious memory, but must bid and compete with other producer agents for the privilege. The middle tier are transporter agents, charged with the task of moving the information from long-term to short-term memory along neural pathways. Finally, the bottom tier are consumer agents that correspond to locations in short-term memory, which must choose some combination of producer and transporter agents to display the desired information at their location.} 
\label{soi}
\end{figure}

In terms of the UDM framework described in Section~\ref{udm} \citep{sm:udm}, the set of elements in this decision object can be represented as $(A, (\Omega, {\cal B}, P), U_\alpha, {\cal F}_\alpha, {\cal I}_\alpha)_{\alpha \in A}$, where $A$ is defined by the set of vertices in this graph representing the decision makers. For example, long-term memory agent  $i$ chooses its actions from the set $U_i$, which can be defined as $\cup_{j,k} Q_{ijk}$. ${\cal F}_i$ is the associated measurable space associated with $U_i$. ${\cal I}_i$ represents the information field of agent $i$, namely its visibility into the decisions made by other entities in the network at the current or past time steps. 

Network economics \citep{nagurney:vibook}is the study of a rich class of equilibrium problems that occur in the real world, from traffic management to supply chains and two-sided online marketplaces. This framework is general, and applies to electronic (e.g., finance) and material (e.g., physical) goods. Here, we are applying the framework to model consciousness.  Each unconscious process has a utility function is defined in terms of the nonnegative service quantity (Q), quality (q), and price ($\pi$) delivered from producer provider $i$ by network provider $j$ to consumer agent $k$.  Production costs, demand functions, delivery costs, and delivery opportunity costs are designated by $f$, $\rho$, $c$, and $oc$ respectively.  Unconscious process  provider $i$ attempts to maximize its utility function $U_i^1(Q,q^*,\pi^*)$ by adjusting $Q_{ijk}$.  Likewise, network provider $j$ attempts to maximize its utility function $U_j^2(Q^*,q,\pi)$ by adjusting $q_{ijk}$ and $\pi_{ijk}$.

\begin{subequations}
\begin{align}
\label{U1}
U_i^1(Q,q^*,\pi^*) &= \sum_{j=1}^n \sum_{k=1}^o \hat{\rho}_{ijk}(Q,q^*)Q_{ijk} - \hat{f}_i(Q)
- \sum_{j=1}^n \sum_{k=1}^o \pi^*_{ijk}Q_{ijk}, \hspace{0.2cm} Q_{ijk} \ge 0 \nonumber
\end{align}
\begin{align}
U_j^2(Q^*,q,\pi) = &\sum_{i=1}^m \sum_{k=1}^o \pi_{ijk}Q^*_{ijk}
- \sum_{i=1}^m \sum_{k=1}^o (c_{ijk}(Q^*,q) + oc_{ijk}(\pi_{ijk})), \nonumber 
q_{ijk}, \pi_{ijk} \ge 0 \nonumber
\end{align}
\end{subequations}

\subsection{Game Theory and Variational Inequalities \index{variational inequalities} }

We now give a brief overview of traditional game theory, and contrast it with the VI framework. Game theory was pioneered by von Neumann and Morgenstern \citep{vonneumann1947}, and later extended by Nash \citep{nash}. An excellent modern overview of game theory is given in \citep{Maschler_Solan_Zamir_2013}. A finite, $n$-person normal form game \index{normal form game} is a tuple $(N,A,U)$, where $N$ is a finite set of $n$ players indexed by $i$, $A = A_1 \times \cdots \times A_n$ is the joint action space formed from the set actions available to each player ($A_i$), and $U$ is a tuple of the players' utility functions (or payoffs) $u_i$ where $u_i: A \rightarrow \mathbb{R}$.  The difficulty in computing the equilibrium depends on the constraints placed on the game.  For instance, two-player, zero-sum games, ensure that player interests are diametrically opposed and can thus be formulated as linear programs (LPs) by the minmax theorem and solved in polynomial time \citep{nisan07}.

By contrast, in two-player, \textit{general}-sum games, \index{general sum game} any increase in one player's utility does not necessarily result in a decrease in the other player's utility so a convenient LP formulation is not possible.  Finding Nash equilibria in two-player, general-sum games is thought to be time exponential in the size of the game in the worst case.  It has been shown that every game has at least one Nash equilibrium which delegates the problem to the class PPAD (polynomial parity argument, directed version) originally designated by Papadimitriou \citep{nisan07}.  Although this game type cannot be converted to an LP, it can be formulated as a linear complimentarily problem (LCP).  In crude terms, the LCP can be formed by introducing an additional constraint called a complementarity condition to the combination of constraints that would appear in each agent's LP had it only been a zero-sum game.  Unlike the LP, the LCP is only composed of constraints making it a pure constraint satisfaction problem (CSP).  The most popular game theoretic algorithm for solving these LCPs is the Lemke-Howson algorithm.  This algorithm performs a series of pivot operations that swap out player strategies until all constraints are satisfied.  An alternate approach is to employ heuristics as in the case of the support-enumeration method (SEM) which repeatedly tests whether a Nash equilibrium exists given a pair of actions, or {\em support profile}.  The heuristic used is to favor testing smaller, more balanced support profiles in order to prune larger regions of the action space.

Finally, we encounter \textit{n}-player, general-sum games, in which the complementarity problem previously defined is now nonlinear (NCP).  One common approach is to approximate the solution of the NCP as solving a {\em sequence} of LCPs (SLCP).  This method is typically fast, however, it is not globally convergent.  Another technique is to solve an optimization problem in which the global minima equate to the Nash equilibria of the original problem.  The drawback is that there are local minima that do not correspond to Nash equilibria making global convergence difficult.  

Some games exhibit a characteristic of {\em payoff independence} where a single player's payoff is dependent on only a subset of the other players in the game.  In this case, the reward table indexed by the joint action-space of all players is overly costly prompting a move from the normal form representation of the game to the more compact representation offered by graphical games.  This can often reduce the space of the representation from exponential to polynomial in the number of players.  When the graph is a tree, a commonly used method, NashProp, computes an $\epsilon$-Nash equilibrium with a back and forth sweep over the graph from the leaves to the root.

\subsection{Stochastic Variational Inequalities \index{stochastic variational inequalities} }

To solve a network economy problem, such as the one shown in Figure~\ref{soi}, we now  introduce classical variational inequalities (VIs) \citep{facchinei-pang:vi}, and outline a metric coinduction type algorithm for stochastic VIs based on \citep{iusem,DBLP:journals/mp/WangB15}. VIs generalize both classical game theory as well as optimization problems, in that any (convex) optimization problem or any Nash game can be formulated as a VI, but the converse is not true. More precisely, a variational inequality model $\cal{M} =$ VI($F,K$), where $F$ is a collection of modular vector-valued functions defined as $F_i$, where $F_i: K_i \subset \mathbb{R}^{n_i} \rightarrow \mathbb{R}^{n_i}$, with each $K_i$ being a convex domain such that $\prod_i K_i = K$. 

\begin{definition}
The category ${\cal C}_{\mbox{VI}}$ of  VIs \index{category of VIs} is defined as one where each object is defined  as a 
finite-dimensional  variational inequality problem ${\cal M}$ = VI($F, K)$, where the vector-valued mapping $F: K \rightarrow \mathbb{R}^n$ is a given continuous function, $K$ is a given closed convex set, and $\langle .,.\rangle$ is the standard inner product in $\mathbb{R}^n$, and the morphisms from one object to another correspond to non-expansive functions.  Solving a  VI is defined as finding a vector $x^* = (x^*_1, \ldots, x^*_n) \in K \subset \mathbb{R}^n$ such that
\begin{equation*}
\langle F(x^*), (y - x^*) \rangle \geq 0, \ \forall y \in K
\end{equation*}
\end{definition}
\begin{figure}[t]
\begin{center}
\begin{minipage}[t]{0.45\textwidth}
\includegraphics[width=\textwidth,height=1.25in]{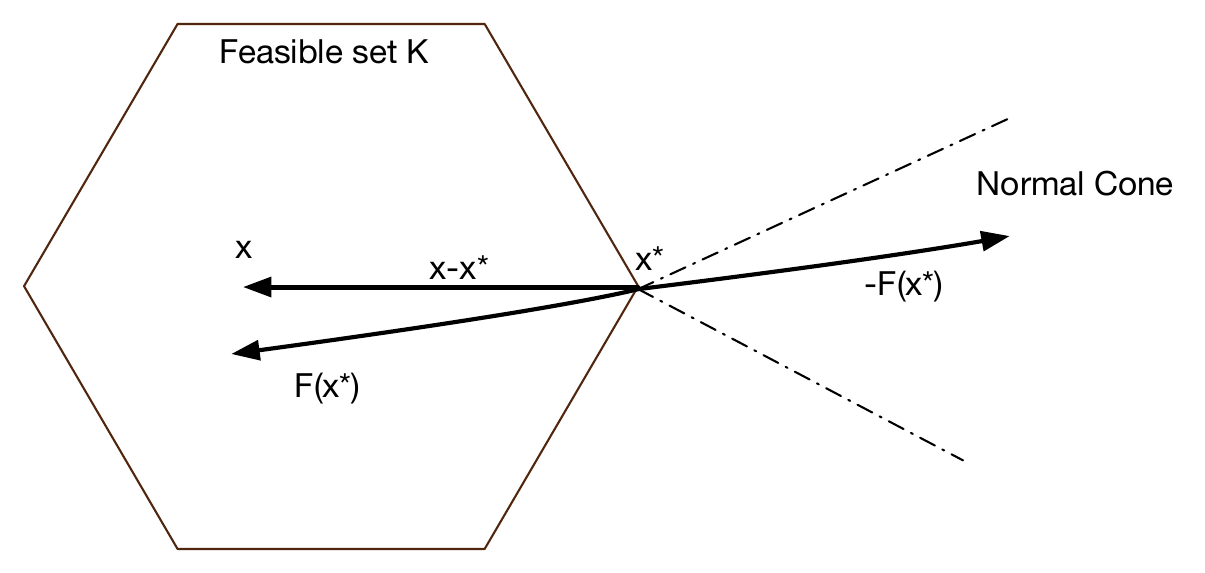}
\end{minipage}
\end{center}
\caption{This figure provides a geometric interpretation of a (deterministic) variational inequality $VI(F,K)$. The mapping $F$ defines a vector field over the feasible set $K$. At the solution point $x^*$, the vector field $F(x^*)$ defines a fixed point, that it, it is directed inwards at the boundary, and  $-F(x^*)$ is an element of the normal cone $C(x^*)$ of $K$ at $x^*$ where the normal cone  $C(x^*)$ at the vector $x^*$ of a convex set $K$ is defined as $C(x^*) = \{y \in \mathbb{R}^n | \langle y, x - x^* \rangle \leq 0, \forall x \in K \}$.}
\end{figure}

We can also define a category of {\em stochastic} VIs as follows. In the stochastic case, we start with a measurable space $({\cal M}, {\cal G})$, a measurable random operator $F: {\cal M} \times \mathbb{R}^n \rightarrow \mathbb{R}^n$, and a random variable $v: \Omega \rightarrow {\cal M}$ defined on a probability space $ (\Omega, {\cal F}, \mathbb{P})$, which enables defining the usual notions of expectation and distributions $P_v$ of $v$. 

\begin{definition}
The category ${\cal C}_{\mbox{SVI}}$ of  stochastic VIs \index{category of stochastic VIs} is defined as one where each object is defined  as a  finite-dimensional  stochastic variational inequality problem ${\cal M}$ = SVI($F, K)$, where the vector-valued mapping $F: K \rightarrow \mathbb{R}^n$ is given by $T(x) = E[F(\xi, x)]$ for all $x \in \mathbb{R}^n$. 
\end{definition}

The solution to a VI is unique, under the following conditions, where $K$ is compact and $F$ is continuous. 
\begin{definition}
$F(x)$ is {\em monotone} \index{monotone VI} if $\langle F(x) - F(y), x - y \rangle \ge 0$, $\forall x, y \in K$.
\end{definition}

\begin{definition}
$F(x)$ is {\em strongly monotone} \index{strongly monotone VI} if $\langle F(x) - F(y), x  - y \rangle \geq \mu \| x - y \|^2_2, \mu > 0, \forall x,y \in K$.
\end{definition}
\begin{definition}
$F(x)$ is {\em Lipschitz} \index{Lipschitz VI} if $\| F(x) - F(y) \|_2 \leq L \|x - y \|_2, \forall x,y \in K$.
\end{definition}

\subsection{An Asynchronous Distributed Algorithm  for solving VIs  }
\label{url-vi} 

 We now discuss an asynchronous distributed algorithm for solving VI's, following the framework introduced in Section~\ref{udm}. There are a wealth of existing methods for deterministic VI's \citep{facchinei-pang:vi,nagurney:vibook}), but these are all batch algorithms.  We describe an adaptation of a stochastic approximation method \citep{rm} to solve VIs.  We now describe an incremental two-step projection method for solving (stochastic) VI's, which is loosely based on \citep{DBLP:journals/mp/WangB15,iusem}. The general update equation can be written as: 
\begin{equation}
\label{2stepalg}
z_k = x_k - \alpha_k F_w(x_k, v_k), \ \ \ x_{k+1} = z_k - \beta_k (z_k - P_{w_k} z_k)
\end{equation}
where $\{v_k\}$ and $\{w_k\}$ are sequences of random variables, generated by sampling a VI model, and $\{\alpha_k\}$ and $\{\beta_k\}$ are sequences of positive scalar step sizes. Note that an interesting feature of this algorithm is that the sequence of iterates $x_k$ is not guaranteed to remain within the feasible space $K$ at each iterate. Indeed, $P_{w_k}$ represents the projection onto a randomly sampled constraint $w_k$. 

The analysis of convergence of this algorithm is somewhat intricate, and we will give the broad highlights as it applies to stochastic VI's. Define the set of random variables ${\cal F}_k = \{v_0, \ldots, v_{k-1}, w_0, \ldots, w_{k-1}, z_0, \ldots, z_{k-1}, x_0, \ldots, x_k \}$. Similar to the convergence of the projection method, it is possible to show that the error of each iteration is {\em stochastically contractive}, in the following sense: 
\[ E[ \| x_{k+1} - x^* \|^2 | {\cal F}_k ] \leq (1 - 2 \mu_k \alpha_k + \delta_k) \| x_k - x^* \|^2 + \epsilon_k, \ \ \ \mbox{w. p. 1} \]
where $\delta_k, \epsilon_k$ are positive errors such that $\sum_{k=0}^\infty \delta_k < \infty$ and $\sum_{k=0}^\infty \epsilon_k < \infty$. Note that the assumption of stochastic contraction is essentially what makes this algorithm an example of metric coinduction \citep{kozen}. 
The convergence of this method rests on the following principal assumptions, stated below: 

\begin{assumption}
The mapping $F_w$ is strongly monotone, and the sampled mapping $F_w(.,v)$ is {\em stochastically Lipschitz continuous} with constant $L > 0$, namely:
\begin{equation}
 E[ \| F_w(x, v_k) - F_w(y, v_k) \|^2 | {\cal F}_k] \leq L^2 \| x - y \|^2. \forall x, y \in \mathbb{R}^n
 \end{equation}
\end{assumption}

\begin{assumption}
The  mapping $F_w$ is bounded, with constant $B > 0$ such that 
\begin{equation}
\| F_w(x^*) \| \leq B, \ \ \ E[\| F_w(x^*, v) \|^2 | {\cal F}_k] \leq B^2, \ \ \forall k \geq 0
\end{equation}
\end{assumption}

\begin{assumption}
The distance between each iterate $x_k$ and the feasible space $K$ reduces ``on average", namely: 
\begin{equation}
\| x - P_K \|^2 \geq \eta \max_{i \in M} \|x - P_{K_i} x\|^2
\end{equation}
where $\eta > 0$ and $M = \{1, \ldots, m \}$ is a finite set of indices such that $\prod K_i = K$. 
\end{assumption}
\begin{assumption}
\[ \sum_{k=0}^\infty \alpha_k = \infty, \sum_{k=0}^\infty \alpha_k^2 < \infty, \sum_{k=0}^\infty \frac{\alpha_k^2}{\gamma_k} < \infty \]
where $\gamma_k = \beta_k (2 - \beta_k)$. 
\end{assumption}

\begin{assumption}
Supermartingale convergence theorem: \index{Supermartingale convergence theorem} Let ${\cal G}_k$ denote the collection of nonnegative random variables $\{y_k\}, \{u_k \}, \{a_k\}, \{b_k\}$ from $i=0, \ldots, k$
\begin{equation}
E[y_{k+1} | {\cal G}_k] \leq (1 + a_k) y_k - u_k + b_k, \ \ \forall k \geq 0, \ \mbox{w.p. 1}
\end{equation}
and $\sum_{k=0}^\infty a_k < \infty$ and $\sum_{k=0}^\infty b_k < \infty$ w.p. 1. Then, $y_k$ converges to a nonnegative random variable, and $\sum_{k=0}^\infty u_k < \infty$.
\end{assumption}
\begin{assumption}
The random variables $w_k, k=0, \ldots$ are such that for $\rho \in (0, 1]$
\[ \inf_{k\geq 0} P(w_k = X_i | {\cal F}_k) \geq \frac{\rho}{m}, \ \ i=1, \ldots, m, \ \mbox{w.p.1} \]
namely, the individual constraints will be sampled sufficiently. Also, the sampling of the stochastic components $v_k, k=0, \ldots$ ensures that
\[ E[F_w(x_k, v_k) | {\cal F}_k] = F_w(x), \ \ \forall x \in \mathbb{R}^n, \ k \geq 0 \]
\end{assumption}

Given the above assumptions, it can be shown that two-step stochastic algorithm given in Equation~\ref{2stepalg} converges to the solution of a (stochastic) VI. 
\begin{theorem}
Given a finite-dimensional stochastic  variational inequality problem is defined by a model ${\cal M}$ = SVI($F, K)$,  where $F(x) = E[F(x, \eta]$, where $E[.]$ now denotes expectation, the two-step algorithm given by Equation~\ref{2stepalg} produces a sequence of iterates $x_k$ that converges almost surely to $x^*$, where 
\begin{equation*}
\langle F_w(x^*), (y - x^*) \rangle \geq 0, \ \forall y \in K
\end{equation*}
\end{theorem}
{\bf Proof:} The proof of this theorem largely follows the derivation given in \citep{iusem,DBLP:journals/mp/WangB15}. $\qed$

To summarize this section, we introduced a network economic framework to model the transmission of information from the large pool of unconscious processes, modeled as producer agents, along neural pathways managed by transporter agents, to locations in short-term memory managed by consumer agents. We described the mathematical framework of variational inequalities (VIs), and described an asynchronous distributed algorithm for solving VIs. This process of moving information from long-term memory into short-term memory can be viewed as a type of ``inverse" URL process described in the previous Section~\ref{udm}.

\section{Summary and Future Work}
\label{summary} 

We described a novel theory of consciousness as a {\em functor} (CF) that receives and transmits contents from unconscious memory into conscious memory. Our CF framework can be seen as a categorial generalization of the Conscious Turing Machine model (CTM) of Blum and Blum, as well as Baars' Global Workspace Theory (GWT).  CF models the ensemble of unconscious processes as a topos category of coalgebras.  As every topos has an internal language defined by a Mitchell-B'enabou language with a Kripke-Joyal semantics, CF is based on an internal ``language of thought" that can be viewed as a categorial version of the ``Brainish" language hypothesized in a CTM . We defined a Multi-modal Universal Mitchell-B'enabou Language Embedding (MUMBLE) that defines the internal Brainish language in our CF framework.  We modeled the transmission of information from conscious short-term working memory to long-term unconscious memory using our recently proposed Universal Reinforcement Learning (URL) framework. To model the transmission of information from unconscious long-term memory into short-term memory, we propose a network economic model, where ``producer" agents correspond to unconscious processes, ``transporter" agents correspond to neural pathways from long-term to short-term memory, and ``consumer agents" correspond to short-term memory locations that use a competitive bidding process to manage the competition between unconscious long-term memory processes. Both URL and the network economic model of consciousness build on a formal theoretical framework for asynchronous parallel distributed computation without the need for synchronization by a global clock.  

Clearly, much remains to be done in elaborating this framework for consciousness in future work. We sketch a few directions that seem worthwhile exploring in the near term. 

\begin{itemize}
    \item {\em Integrating short-term and long-term memory through adjoint functors:} We sketched out the elements of our CF framework in terms of a set of building blocks, but further elaboration of how exactly these pieces are assembled to realize consciousness is needed. Fundamentally, we posited that unconscious processes are modeled as a topos of coalgebras, and that long-term unconscious behavior resulting from a (universal) reinforcement learning process results in creating a topos structure of (action)value functions. We used a network economic model to define the problem of placing information into resource-constrained short-term memory from long-term memory through a complex network game, solved using an asynchronous VI method. We believe a deeper integration of URL and ``inverse URL" is possible. 

    \item {\em Computational implementation of CF: } We rightly ignored the important issue of how to implement our CF framework in this initial introductory paper, but it is worthwhile to emphasize that any computational theory must ultimately be tested on the crucible of actual software implementation. We believe the CF framework is implementable using any of the standard deep learning or LLM tools available currently, suitably generalized to our categorical setting. 

    \item {\em Neural plausibility:} We remarked that in order for the brain to be capable of realizing causal, probabilistic and statistical computation, it must be fundamentally capable of multiplying, copying, and deleting objects. This result comes from basic advances in categorical probability \citep{Fritz_2020}, and it is an intriguing question to ask how the brain manages to neurally implement these operations. We are not the first to hypothesize that the brain must have such capabilities. \citet{gallistel-king} remark that ``there must be an addressable read-write memory mechanism in the brains". Our CF model may shed deeper light on why this hypothesis is more plausible. 
\end{itemize}


\newpage 

 \begin{thebibliography}{82}
\providecommand{\natexlab}[1]{#1}
\providecommand{\url}[1]{\texttt{#1}}
\expandafter\ifx\csname urlstyle\endcsname\relax
  \providecommand{\doi}[1]{doi: #1}\else
  \providecommand{\doi}{doi: \begingroup \urlstyle{rm}\Url}\fi

\bibitem[Aczel(1988)]{Aczel1988-ACZNS}
Peter Aczel.
\newblock \emph{Non-Well-Founded Sets}.
\newblock {CSLI{}} Lecture Notes, Palo Alto, CA, USA, 1988.

\bibitem[Aczel and Mendler(1989)]{aczel-final-coalgebra-thm}
Peter Aczel and Nax Mendler.
\newblock A final coalgebra theorem.
\newblock In David~H. Pitt, David~E. Rydeheard, Peter Dybjer, Andrew~M. Pitts, and Axel Poign{\'e}, editors, \emph{Category Theory and Computer Science}, pages 357--365, Berlin, Heidelberg, 1989. Springer Berlin Heidelberg.
\newblock ISBN 978-3-540-46740-3.

\bibitem[Alfredo N.~Iusem(2018)]{iusem}
Philip~Thompson Alfredo N.~Iusem, Alejandro~Jofré.
\newblock Incremental constraint projection methods for monotone stochastic variational inequalities.
\newblock \emph{Mathematics of Operations Research}, 44\penalty0 (1):\penalty0 236--263, 2018.
\newblock URL \url{https://doi.org/10.1287/moor.2017.0922}.

\bibitem[Arbib and Manes(1974)]{DBLP:conf/category/ArbibM74a}
Michael~A. Arbib and Ernest~G. Manes.
\newblock A categorist's view of automata and systems.
\newblock In Ernest~G. Manes, editor, \emph{Category Theory Applied to Computation and Control, Proceedings of the First International Symposium, San Francisco, CA, USA, February 25-26, 1974, Proceedings}, volume~25 of \emph{Lecture Notes in Computer Science}, pages 51--64. Springer, 1974.
\newblock \doi{10.1007/3-540-07142-3\_61}.
\newblock URL \url{https://doi.org/10.1007/3-540-07142-3\_61}.

\bibitem[Arora and Barak(2009)]{arora}
Sanjeev Arora and Boaz Barak.
\newblock \emph{Computational Complexity - {A} Modern Approach}.
\newblock Cambridge University Press, 2009.
\newblock ISBN 978-0-521-42426-4.
\newblock URL \url{http://www.cambridge.org/catalogue/catalogue.asp?isbn=9780521424264}.

\bibitem[Baars(1997{\natexlab{a}})]{baars:oup}
Bernard Baars.
\newblock \emph{In the theater of consciousness: The workspace of the mind}.
\newblock Oxford Univ. Press, New York, NY, 1997{\natexlab{a}}.
\newblock URL \url{https://psycnet.apa.org/doi/10.1093/acprof:oso/9780195102659.001.1}.

\bibitem[Baars(1997{\natexlab{b}})]{baars:theater}
Bernard Baars.
\newblock \emph{In the Theater of Consciousness: The Workspace of the Mind}.
\newblock Oxford University Press, 1997{\natexlab{b}}.

\bibitem[Bell(1988)]{bell}
J.~L. Bell.
\newblock \emph{Toposes and Local Set Theories}.
\newblock Dover, 1988.

\bibitem[Bengio(2019)]{bengio2019consciousnessprior}
Yoshua Bengio.
\newblock The consciousness prior, 2019.
\newblock URL \url{https://arxiv.org/abs/1709.08568}.

\bibitem[Bertsekas(2019)]{bertsekas:rlbook}
Dimitri Bertsekas.
\newblock \emph{Reinforcement Learning and Optimal Control}.
\newblock Athena Scientific, 2019.

\bibitem[Bertsekas(2005)]{DBLP:books/lib/Bertsekas05}
Dimitri~P. Bertsekas.
\newblock \emph{Dynamic programming and optimal control, 3rd Edition}.
\newblock Athena Scientific, 2005.
\newblock ISBN 1886529264.
\newblock URL \url{https://www.worldcat.org/oclc/314894080}.

\bibitem[Bertsekas and Tsitsiklis(1997)]{bertsekas:pdc}
Dimitri~P. Bertsekas and John~N. Tsitsiklis.
\newblock \emph{Parallel and Distributed Computation: Numerical Methods}.
\newblock Athena Scientific, 1997.
\newblock ISBN 1886529019.

\bibitem[Blum and Blum(2022)]{blum:pnas}
Lenore Blum and Manuel Blum.
\newblock A theory of consciousness from a theoretical computer science perspective: Insights from the conscious turing machine.
\newblock \emph{Proceedings of the National Academy of Sciences}, 119\penalty0 (21):\penalty0 e2115934119, 2022.
\newblock \doi{10.1073/pnas.2115934119}.
\newblock URL \url{https://www.pnas.org/doi/abs/10.1073/pnas.2115934119}.

\bibitem[Blum and Blum(2021)]{blum2021theoreticalcomputerscienceperspective}
Manuel Blum and Lenore Blum.
\newblock A theoretical computer science perspective on consciousness, 2021.
\newblock URL \url{https://arxiv.org/abs/2011.09850}.

\bibitem[Bradley et~al.(2022)Bradley, Terilla, and Vlassopoulos]{bradley:enriched-yoneda-llms}
TD. Bradley, J.~Terilla, and Y.~Vlassopoulos.
\newblock An enriched category theory of language: From syntax to semantics.
\newblock \emph{La Matematica}, 1:\penalty0 551--580, 2022.

\bibitem[Carlsson and Memoli(2010)]{Carlsson2010}
Gunnar Carlsson and Facundo Memoli.
\newblock Classifying clustering schemes, 2010.
\newblock URL \url{http://arxiv.org/abs/1011.5270}.
\newblock cite arxiv:1011.5270.

\bibitem[Carpentier et~al.(2015)Carpentier, Chancelier, Cohen, and De~Lara]{carpentier2015stochastic}
P.~Carpentier, J.P. Chancelier, G.~Cohen, and M.~De~Lara.
\newblock \emph{Stochastic Multi-Stage Optimization: At the Crossroads Between Discrete Time Stochastic Control and Stochastic Programming}.
\newblock Probability theory and stochastic modelling. Springer International Publishing, 2015.
\newblock ISBN 9783319181394.
\newblock URL \url{https://books.google.com/books?id=w5PizQEACAAJ}.

\bibitem[Chalmers(1996{\natexlab{a}})]{chalmers}
D.~Chalmers.
\newblock \emph{The Conscious Mind}.
\newblock Oxford University Press, 1996{\natexlab{a}}.

\bibitem[Chalmers(1996{\natexlab{b}})]{chalmers:theory}
David Chalmers.
\newblock \emph{The Conscious Mind: In Search of a Fundamental Theory}.
\newblock Oxford, 1996{\natexlab{b}}.

\bibitem[Cho and Jacobs(2019)]{Cho_2019}
Kenta Cho and Bart Jacobs.
\newblock Disintegration and bayesian inversion via string diagrams.
\newblock \emph{Mathematical Structures in Computer Science}, 29\penalty0 (7):\penalty0 938–971, March 2019.
\newblock ISSN 1469-8072.
\newblock \doi{10.1017/s0960129518000488}.
\newblock URL \url{http://dx.doi.org/10.1017/S0960129518000488}.

\bibitem[Chomsky(1959)]{chomsky59}
Noam Chomsky.
\newblock On certain formal properties of grammars.
\newblock \emph{Information and Control}, 2\penalty0 (2):\penalty0 137--167, June 1959.
\newblock URL \url{http://www.diku.dk/hjemmesider/ansatte/henglein/papers/chomsky1959.pdf}.

\bibitem[Chomsky(1965)]{chomsky1965}
Noam Chomsky.
\newblock \emph{Aspects of the Theory of Syntax}.
\newblock The MIT Press, Cambridge, 1965.
\newblock URL \url{http://www.amazon.com/Aspects-Theory-Syntax-Noam-Chomsky/dp/0262530074}.

\bibitem[Crick and Koch(1990)]{Crick1990-CRITAN}
Francis Crick and Christof Koch.
\newblock Toward a neurobiological theory of consciousness.
\newblock \emph{Seminars in the Neurosciences}, 2:\penalty0 263--275, 1990.

\bibitem[Dehaene(2014)]{dehaene}
Stanis;as Dehaene.
\newblock \emph{Cpnsciousness and the Brain}.
\newblock Penguin Books, 2014.

\bibitem[Dennett(1991)]{dennett}
D.~C. Dennett.
\newblock \emph{Consciousness Explained}.
\newblock Little, Brown, and Company, 1991.

\bibitem[Descartes(1644)]{descartes}
Rene Descartes.
\newblock \emph{Principles of Philosophy}.
\newblock Dordecht: Reidel, 1644.

\bibitem[Edelman and Tononi(2000)]{edelman:consciousness}
Gerard Edelman and Giulio Tononi.
\newblock \emph{A Universe of Consciousness: How Matter becomes Imagination}.
\newblock Basic Books, 2000.

\bibitem[Facchinei and Pang(2003)]{facchinei-pang:vi}
F.~Facchinei and J.~Pang.
\newblock \emph{Finite-Dimensional Variational Inequalities and Complementarity Problems}.
\newblock Springer, 2003.

\bibitem[Feys et~al.(2018)Feys, Hansen, and Moss]{feys}
Frank M.~V. Feys, Helle~Hvid Hansen, and Lawrence~S. Moss.
\newblock Long-term values in markov decision processes, (co)algebraically.
\newblock In \emph{Coalgebraic Methods in Computer Science: 14th IFIP WG 1.3 International Workshop, CMCS 2018, Colocated with ETAPS 2018, Thessaloniki, Greece, April 14–15, 2018, Revised Selected Papers}, page 78–99, Berlin, Heidelberg, 2018. Springer-Verlag.
\newblock ISBN 978-3-030-00388-3.
\newblock \doi{10.1007/978-3-030-00389-0_6}.
\newblock URL \url{https://doi.org/10.1007/978-3-030-00389-0_6}.

\bibitem[Fodor(1980)]{fodor:loth}
Jerry Fodor.
\newblock \emph{The Language of Thought}.
\newblock Harvard University Press, USA, 1980.
\newblock ISBN 9780674510302.

\bibitem[Fong and Spivak(2018)]{fong2018seven}
Brendan Fong and David~I Spivak.
\newblock \emph{Seven Sketches in Compositionality: An Invitation to Applied Category Theory}.
\newblock Cambridge University Press, 2018.

\bibitem[Fox(1976)]{fox}
James Fox.
\newblock \emph{Universal Coalgebras}.
\newblock Dissertation, McGill University, 1976.

\bibitem[Fritz(2020)]{Fritz_2020}
Tobias Fritz.
\newblock A synthetic approach to markov kernels, conditional independence and theorems on sufficient statistics.
\newblock \emph{Advances in Mathematics}, 370:\penalty0 107239, August 2020.
\newblock ISSN 0001-8708.
\newblock \doi{10.1016/j.aim.2020.107239}.
\newblock URL \url{http://dx.doi.org/10.1016/j.aim.2020.107239}.

\bibitem[Fritz and Klingler(2023)]{fritz:jmlr}
Tobias Fritz and Andreas Klingler.
\newblock The d-separation criterion in categorical probability.
\newblock \emph{Journal of Machine Learning Research}, 24\penalty0 (46):\penalty0 1--49, 2023.
\newblock URL \url{http://jmlr.org/papers/v24/22-0916.html}.

\bibitem[Gallistel and King(2010)]{gallistel-king}
C.R. Gallistel and Adam~Philip King.
\newblock \emph{Memory and the Computational Brain}.
\newblock Wiley-Blackwell, United States, 2010.

\bibitem[Goldblatt(2006)]{goldblatt:topos}
Robert Goldblatt.
\newblock \emph{Topoi: The Categorial Analysis of Logic}.
\newblock Dover Press, 2006.

\bibitem[Heunen and Vicary(2019)]{Heunen2019}
Chris Heunen and Jamie Vicary.
\newblock \emph{Categories for Quantum Theory: An Introduction}.
\newblock Oxford University Press, November 2019.
\newblock \doi{10.1093/oso/9780198739623.001.0001}.

\bibitem[Jacobs(2016)]{jacobs:book}
Bart Jacobs.
\newblock \emph{Introduction to Coalgebra: Towards Mathematics of States and Observation}, volume~59 of \emph{Cambridge Tracts in Theoretical Computer Science}.
\newblock Cambridge University Press, 2016.
\newblock ISBN 9781316823187.
\newblock \doi{10.1017/CBO9781316823187}.
\newblock URL \url{https://doi.org/10.1017/CBO9781316823187}.

\bibitem[Jacobs et~al.(2021)Jacobs, Kissinger, and Zanasi]{DBLP:journals/mscs/JacobsKZ21}
Bart Jacobs, Aleks Kissinger, and Fabio Zanasi.
\newblock Causal inference via string diagram surgery: {A} diagrammatic approach to interventions and counterfactuals.
\newblock \emph{Math. Struct. Comput. Sci.}, 31\penalty0 (5):\penalty0 553--574, 2021.
\newblock \doi{10.1017/S096012952100027X}.
\newblock URL \url{https://doi.org/10.1017/S096012952100027X}.

\bibitem[Johnstone(2002)]{Johnstone:592033}
Peter~T Johnstone.
\newblock \emph{{Sketches of an elephant: a Topos theory compendium}}.
\newblock Oxford logic guides. Oxford Univ. Press, New York, NY, 2002.
\newblock URL \url{https://cds.cern.ch/record/592033}.

\bibitem[Johnstone(2014)]{Johnstone:topostheory}
Peter~T Johnstone.
\newblock \emph{{Topos Theory}}.
\newblock Dover Publications, 2014.

\bibitem[Kahneman(2011)]{kahneman2011thinking}
Daniel Kahneman.
\newblock \emph{Thinking, fast and slow}.
\newblock Farrar, Straus and Giroux, New York, 2011.
\newblock ISBN 9780374275631 0374275637.
\newblock URL \url{https://www.amazon.de/Thinking-Fast-Slow-Daniel-Kahneman/dp/0374275637/ref=wl_it_dp_o_pdT1_nS_nC?ie=UTF8&colid=151193SNGKJT9&coliid=I3OCESLZCVDFL7}.

\bibitem[Kozen and Ruozzi(2009)]{kozen}
Dexter Kozen and Nicholas Ruozzi.
\newblock Applications of metric coinduction.
\newblock \emph{Log. Methods Comput. Sci.}, 5\penalty0 (3), 2009.
\newblock URL \url{http://arxiv.org/abs/0908.2793}.

\bibitem[Locke(1688)]{locke}
J.~Locke.
\newblock \emph{John Locke: An Essay concerning Human Understanding}.
\newblock Oxford World's Classics. Oxford University Press, United States, 1688.
\newblock ISBN 978-0-19-929662-0.

\bibitem[MacLane(1971)]{maclane:71}
Saunders MacLane.
\newblock \emph{Categories for the Working Mathematician}.
\newblock Springer-Verlag, New York, 1971.
\newblock Graduate Texts in Mathematics, Vol. 5.

\bibitem[MacLane and leke Moerdijk(1994)]{maclane:sheaves}
Saunders MacLane and leke Moerdijk.
\newblock \emph{Sheaves in Geometry and Logic: A First Introduction to Topos Theory}.
\newblock Springer, 1994.

\bibitem[Mahadevan(2021{\natexlab{a}})]{sm:udm}
Sridhar Mahadevan.
\newblock Universal decision models.
\newblock \emph{CoRR}, abs/2110.15431, 2021{\natexlab{a}}.
\newblock URL \url{https://arxiv.org/abs/2110.15431}.

\bibitem[Mahadevan(2021{\natexlab{b}})]{sm:uig}
Sridhar Mahadevan.
\newblock Universal imitation games.
\newblock \emph{CoRR}, abs/2110.15431, 2021{\natexlab{b}}.
\newblock URL \url{https://arxiv.org/abs/2110.15431}.

\bibitem[Mahadevan(2023)]{DBLP:journals/entropy/Mahadevan23}
Sridhar Mahadevan.
\newblock Universal causality.
\newblock \emph{Entropy}, 25\penalty0 (4):\penalty0 574, 2023.
\newblock \doi{10.3390/E25040574}.
\newblock URL \url{https://doi.org/10.3390/e25040574}.

\bibitem[Mahadevan(2024)]{mahadevan2024gaiacategoricalfoundationsgenerative}
Sridhar Mahadevan.
\newblock Gaia: Categorical foundations of generative ai, 2024.
\newblock URL \url{https://arxiv.org/abs/2402.18732}.

\bibitem[Mahadevan(2025{\natexlab{a}})]{cktheory}
Sridhar Mahadevan.
\newblock A higher algebraic k-theory of causality.
\newblock \emph{Entropy}, In Press 2025{\natexlab{a}}.
\newblock \doi{10.20944/preprints202501.1242.v1}.
\newblock URL \url{https://doi.org/10.20944/preprints202501.1242.v1}.

\bibitem[Mahadevan(2025{\natexlab{b}})]{mahadevan2025toposcausalmodels}
Sridhar Mahadevan.
\newblock Topos causal models, 2025{\natexlab{b}}.
\newblock URL \url{https://arxiv.org/abs/2508.08295}.

\bibitem[Mahadevan(2025{\natexlab{c}})]{mahadevan2025topostheorygenerativeai}
Sridhar Mahadevan.
\newblock Topos theory for generative ai and llms, 2025{\natexlab{c}}.
\newblock URL \url{https://arxiv.org/abs/2508.08293}.

\bibitem[Mahadevan(2025{\natexlab{d}})]{mahadevan2025universalreinforcementlearningcoalgebras}
Sridhar Mahadevan.
\newblock Universal reinforcement learning in coalgebras: Asynchronous stochastic computation via conduction, 2025{\natexlab{d}}.
\newblock URL \url{https://arxiv.org/abs/2508.15128}.

\bibitem[Mahadevan(2025{\natexlab{e}})]{sm:aig}
Sridhar Mahadevan.
\newblock \emph{Artificial General Intelligence: A Categorial Theory}.
\newblock Springer, 2025{\natexlab{e}}.
\newblock In Press.

\bibitem[Marr(1982)]{Marr:1982:VCI:1095712}
David Marr.
\newblock \emph{Vision: A Computational Investigation into the Human Representation and Processing of Visual Information}.
\newblock Henry Holt and Co., Inc., New York, NY, USA, 1982.
\newblock ISBN 0716715678.

\bibitem[Marr(1990)]{marr:personal-view}
David Marr.
\newblock \emph{AI: a personal view}, page 97–107.
\newblock Cambridge University Press, USA, 1990.
\newblock ISBN 0521359449.

\bibitem[Maschler et~al.(2013)Maschler, Solan, and Zamir]{Maschler_Solan_Zamir_2013}
Michael Maschler, Eilon Solan, and Shmuel Zamir.
\newblock \emph{Game Theory}.
\newblock Cambridge University Press, 2013.

\bibitem[McInnes et~al.(2018)McInnes, Healy, and Melville]{umap}
Leland McInnes, John Healy, and James Melville.
\newblock Umap: Uniform manifold approximation and projection for dimension reduction, 2018.
\newblock URL \url{https://arxiv.org/abs/1802.03426}.

\bibitem[Mitchell(1972)]{mitchell:topoi}
William Mitchell.
\newblock Boolean topoi and the theory of sets.
\newblock \emph{LJournal of Pure and Applied Algebra}, 2:\penalty0 261--274, 1972.

\bibitem[Morris et~al.(2023)Morris, Sohl-Dickstein, Fiedel, Warkentin, Dafoe, Faust, Farabet, and Legg]{agi-dm}
Meredith~Ringel Morris, Jascha Sohl-Dickstein, Noah Fiedel, Tris Warkentin, Allan Dafoe, Aleksandra Faust, Clement Farabet, and Shane Legg, editors.
\newblock \emph{Levels of AGI for Operationalizing Progress on the Path to AGI}, 2023.
\newblock Original arXiv title in November 2023 was "Levels of AGI": Operationalizing Progress on the Path to AGI. Final title for publication as a position paper at ICML 2024 is: Levels of AGI for Operataionalizing Progress on the Path to AGI.

\bibitem[Nagurney(1999)]{nagurney:vibook}
A.~Nagurney.
\newblock \emph{Network Economics: A Variational Inequality Approach}.
\newblock Kluwer Academic Press, 1999.

\bibitem[Nash(1951)]{nash}
John Nash.
\newblock Non-cooperative games.
\newblock \emph{Annals of Mathematics}, 54\penalty0 (2):\penalty0 286--295, 1951.
\newblock URL \url{https://doi.org/10.2307/1969529}.

\bibitem[Nisan et~al.(2007)Nisan, Roughgarden, Tardos, and Vazirani]{nisan07}
Noam Nisan, Tim Roughgarden, Eva Tardos, and Vijay Vazirani.
\newblock \emph{Algorithmic Game Theory}.
\newblock Cambridge University Press, Cambridge; New York, 2007.
\newblock ISBN 9780521872829 0521872820.

\bibitem[Pearl(2009)]{pearl-book}
Judea Pearl.
\newblock \emph{Causality: Models, Reasoning and Inference}.
\newblock Cambridge University Press, USA, 2nd edition, 2009.
\newblock ISBN 052189560X.

\bibitem[Puterman(1990)]{PUTERMAN1990331}
Martin~L. Puterman.
\newblock Chapter 8 markov decision processes.
\newblock In \emph{Stochastic Models}, volume~2 of \emph{Handbooks in Operations Research and Management Science}, pages 331--434. Elsevier, 1990.
\newblock \doi{https://doi.org/10.1016/S0927-0507(05)80172-0}.
\newblock URL \url{https://www.sciencedirect.com/science/article/pii/S0927050705801720}.

\bibitem[Richter(2020)]{richter2020categories}
B.~Richter.
\newblock \emph{From Categories to Homotopy Theory}.
\newblock Cambridge Studies in Advanced Mathematics. Cambridge University Press, 2020.
\newblock ISBN 9781108479622.
\newblock URL \url{https://books.google.com/books?id=pnzUDwAAQBAJ}.

\bibitem[Riehl(2017)]{riehl2017category}
E.~Riehl.
\newblock \emph{Category Theory in Context}.
\newblock Aurora: Dover Modern Math Originals. Dover Publications, 2017.
\newblock ISBN 9780486820804.
\newblock URL \url{https://books.google.com/books?id=6B9MDgAAQBAJ}.

\bibitem[Robbins and Monro(1951)]{rm}
Herbert Robbins and Sutton Monro.
\newblock {A Stochastic Approximation Method}.
\newblock \emph{The Annals of Mathematical Statistics}, 22\penalty0 (3):\penalty0 400 -- 407, 1951.
\newblock \doi{10.1214/aoms/1177729586}.
\newblock URL \url{https://doi.org/10.1214/aoms/1177729586}.

\bibitem[Rutten(2005)]{rutten:streams}
J.~J. M.~M. Rutten.
\newblock A coinductive calculus of streams.
\newblock \emph{Mathematical. Structures in Comp. Sci.}, 15\penalty0 (1):\penalty0 93–147, February 2005.
\newblock ISSN 0960-1295.
\newblock \doi{10.1017/S0960129504004517}.
\newblock URL \url{https://doi.org/10.1017/S0960129504004517}.

\bibitem[Rutten(2000)]{rutten2000universal}
J.J.M.M. Rutten.
\newblock Universal coalgebra: a theory of systems.
\newblock \emph{Theoretical Computer Science}, 249\penalty0 (1):\penalty0 3 -- 80, 2000.
\newblock ISSN 0304-3975.
\newblock \doi{http://dx.doi.org/10.1016/S0304-3975(00)00056-6}.
\newblock URL \url{http://www.sciencedirect.com/science/article/pii/S0304397500000566}.
\newblock Modern Algebra.

\bibitem[Searle(1997)]{searle:consciousness}
John~R. Searle.
\newblock \emph{The Mystery of Consciousness}.
\newblock New York Review Book, 1997.

\bibitem[Selinger(2010)]{Selinger_2010}
P.~Selinger.
\newblock A survey of graphical languages for monoidal categories.
\newblock In \emph{New Structures for Physics}, pages 289--355. Springer Berlin Heidelberg, 2010.
\newblock \doi{10.1007/978-3-642-12821-9_4}.
\newblock URL \url{https://doi.org/10.1007%2F978-3-642-12821-9_4}.

\bibitem[Singh et~al.(2004)Singh, James, and Rudary]{singh-uai04}
Satinder~P. Singh, Michael~R. James, and Matthew~R. Rudary.
\newblock Predictive state representations: {A} new theory for modeling dynamical systems.
\newblock In David~Maxwell Chickering and Joseph~Y. Halpern, editors, \emph{{UAI} '04, Proceedings of the 20th Conference in Uncertainty in Artificial Intelligence, Banff, Canada, July 7-11, 2004}, pages 512--518. {AUAI} Press, 2004.
\newblock URL \url{https://dslpitt.org/uai/displayArticleDetails.jsp?mmnu=1\&smnu=2\&article\_id=1148\&proceeding\_id=20}.

\bibitem[Sokolova(2011)]{SOKOLOVA20115095}
Ana Sokolova.
\newblock Probabilistic systems coalgebraically: A survey.
\newblock \emph{Theoretical Computer Science}, 412\penalty0 (38):\penalty0 5095--5110, 2011.
\newblock ISSN 0304-3975.
\newblock \doi{https://doi.org/10.1016/j.tcs.2011.05.008}.
\newblock URL \url{https://www.sciencedirect.com/science/article/pii/S0304397511003902}.
\newblock CMCS Tenth Anniversary Meeting.

\bibitem[Spivak and Kent(2012)]{Spivak_2012}
David~I. Spivak and Robert~E. Kent.
\newblock Ologs: A categorical framework for knowledge representation.
\newblock \emph{{PLoS} {ONE}}, 7\penalty0 (1):\penalty0 e24274, jan 2012.
\newblock \doi{10.1371/journal.pone.0024274}.

\bibitem[Sutton and Barto(2018)]{sb:2018}
Richard~S. Sutton and Andrew~G. Barto.
\newblock \emph{Reinforcement Learning: An Introduction}.
\newblock The MIT Press, second edition, 2018.
\newblock URL \url{http://incompleteideas.net/book/the-book-2nd.html}.

\bibitem[Tsitsiklis(1993)]{tsitsiklis}
J.N. Tsitsiklis.
\newblock Asynchronous stochastic approximation and q-learning.
\newblock In \emph{Proceedings of 32nd IEEE Conference on Decision and Control}, pages 395--400 vol.1, 1993.
\newblock \doi{10.1109/CDC.1993.325119}.

\bibitem[Vigna(2003)]{vigna2003guided}
Sebastiano Vigna.
\newblock A guided tour in the topos of graphs, 2003.

\bibitem[von Neumann and Morgenstern(1947)]{vonneumann1947}
J.~von Neumann and O.~Morgenstern.
\newblock \emph{Theory of games and economic behavior}.
\newblock Princeton University Press, 1947.

\bibitem[Wang and Bertsekas(2015)]{DBLP:journals/mp/WangB15}
Mengdi Wang and Dimitri~P. Bertsekas.
\newblock Incremental constraint projection methods for variational inequalities.
\newblock \emph{Math. Program.}, 150\penalty0 (2):\penalty0 321--363, 2015.
\newblock \doi{10.1007/s10107-014-0769-x}.
\newblock URL \url{https://doi.org/10.1007/s10107-014-0769-x}.

\bibitem[Witsenhausen(1975)]{witsenhausen:1975}
H.~S. Witsenhausen.
\newblock The intrinsic model for discrete stochastic control: Some open problems.
\newblock In A.~Bensoussan and J.~L. Lions, editors, \emph{Control Theory, Numerical Methods and Computer Systems Modelling}, pages 322--335, Berlin, Heidelberg, 1975. Springer Berlin Heidelberg.

\end{thebibliography}
\end{document}